%% file: acl_latex.tex
\pdfoutput=1

\documentclass[11pt]{article}

\usepackage[final]{acl}

\usepackage{times}
\usepackage{latexsym}
\usepackage{titletoc}
\usepackage{tcolorbox}
\usepackage{tabularx}
\usepackage{booktabs}
\tcbuselibrary{skins,breakable}
\usepackage{listings}
\usepackage{listingsutf8}

\usepackage[T1]{fontenc}

\usepackage[utf8]{inputenc}
\usepackage{newunicodechar}
\usepackage{amssymb}
\newunicodechar{✓}{\checkmark}
\newunicodechar{✗}{\ding{55}}

\usepackage{microtype}
\usepackage{enumitem}
\setlist{nosep}

\usepackage{inconsolata}

\usepackage{graphicx}
\usepackage{amsmath} 
\usepackage{amsmath, amssymb} 
\usepackage{tikz}
\usetikzlibrary{positioning} 
\usepackage{hyperref}
\usepackage{multirow}
\usepackage[table,xcdraw]{xcolor}
\usepackage[dvipsnames]{xcolor}
\tcbuselibrary{skins,breakable,listings}
\usepackage{listings}
\usepackage{listingsutf8}
\usepackage{pifont} 
\usepackage{subcaption}
\usepackage{caption}
\usepackage{booktabs}
\usepackage{fontawesome5}
\usepackage[ruled,vlined,linesnumbered]{algorithm2e}
\usepackage{float}

\usepackage{setspace}

\newcommand{\EM}{\mathrm{EM}}
\newcommand{\EX}{\mathrm{EX}}
\newcommand{\Traj}{\mathcal{T}}

\DontPrintSemicolon                    
\SetCommentSty{textnormal}             

\newcommand{\Stage}[1]{%
  \tcp*[l]{\colorbox{gray!20}{\strut \texttt{}\ #1}}%
}

\definecolor{ThoughtBg}{RGB}{245,242,255}   
\definecolor{ThoughtBd}{RGB}{128,90,213}    
\definecolor{ActionBg}{RGB}{255,247,237}    
\definecolor{ActionBd}{RGB}{234,88,12}      
\definecolor{ObsBg}{RGB}{240,253,244}       
\definecolor{ObsBd}{RGB}{22,163,74}         
\definecolor{CodeBg}{gray}{0.96}

\lstdefinelanguage{LLMTags}{
  moredelim=**[is][\color{RoyalPurple}\bfseries\ttfamily]{<}{>}, 
}

\lstdefinestyle{llmmono}{
  language=LLMTags,
  basicstyle=\ttfamily\small,
  columns=fullflexible,
  keepspaces=true,
  showstringspaces=false,
  breaklines=true,
  xleftmargin=0pt,
  xrightmargin=0pt,
  frame=single, framerule=0pt, rulecolor=\color{black!8},
  backgroundcolor=\color{CodeBg},
  inputencoding=utf8,
  literate={→}{{$\to$}}1 {—}{{---}}1 {…}{{\ldots}}1 {“}{{``}}1 {”}{{''}}1 {’}{{'}}1,
}

\tcbset{
  enhanced, breakable, rounded corners,
  coltitle=black, fonttitle=\bfseries\footnotesize,
  boxsep=1mm, left=2mm, right=2mm, top=1.2mm, bottom=1.2mm,
  colframe=black!10, colback=white, borderline={0.4mm}{0mm}{black!8},
  attach boxed title to top left,
  boxed title style={
    size=small, colback=black!4, colframe=black!10, boxsep=1mm,
    rounded corners
  }
}

\newtcblisting{ThoughtBox}{
  title={\faRobot\; Thought},
  colback=ThoughtBg,
  colframe=ThoughtBd,
  borderline={0mm}{0mm}{ThoughtBd},
  left=1.5mm,
right=1.5mm,
boxsep=2pt,  
  listing engine=listings, 
  listing only, 
listing options={
  aboveskip=0pt,
  belowskip=0pt,
  postbreak=\mbox{}, 
  breakindent=0pt,
escapeinside={(*@}{@*)},
  style=llmmono,
  basicstyle=\ttfamily\scriptsize,
  keepspaces=true,    
  breaklines=true,      
  breakautoindent=false, 
  breakindent=0pt,      
  xleftmargin=0pt,
  framexleftmargin=0pt,
  columns=fullflexible,
  showstringspaces=false,
  postbreak=\mbox{},     
   literate={<}{{\textless}}1 {>}{{\textgreater}}1,
}
}

\newtcblisting{ActionBox}{
  title={\faTools\; Tool Calling (Query DataBase / Memory Retrieve and Coherence Verify)}, colback=ActionBg, colframe=ActionBd, borderline={0.6mm}{0mm}{ActionBd},
  left=1.5mm,
right=1.5mm,
boxsep=2pt,  
  listing engine=listings, 
  listing only, 
listing options={
  aboveskip=0pt,
  belowskip=0pt,
  breakindent=0pt,
  postbreak=\mbox{}, 
  style=llmmono,
  basicstyle=\ttfamily\scriptsize,
  keepspaces=true,       
  breaklines=true,       
  breakautoindent=false, 
  breakindent=0pt,       
  xleftmargin=0pt,
  framexleftmargin=0pt,
  columns=fullflexible,
  showstringspaces=false,
  postbreak=\mbox{},      
   literate={<}{{\textless}}1 {>}{{\textgreater}}1,
}
}
\newtcblisting{ObsBox}{
  title={\faEye\; Tool Response}, colback=ObsBg, colframe=ObsBd, borderline={0.6mm}{0mm}{ObsBd},
  left=1.5mm,
right=1.5mm,
boxsep=2pt,  
  listing engine=listings, 
  listing only, 
listing options={
  aboveskip=0pt,
  belowskip=0pt,
  breakindent=0pt,
  postbreak=\mbox{}, 
  style=llmmono,
  basicstyle=\ttfamily\scriptsize,
  keepspaces=true,       
  breaklines=true,       
  breakautoindent=false, 
  breakindent=0pt,       
  xleftmargin=0pt,
  framexleftmargin=0pt,
  columns=fullflexible,
  showstringspaces=false,
  postbreak=\mbox{},      
   literate={<}{{\textless}}1 {>}{{\textgreater}}1,
}
}

\newtcblisting{prompt}{
  title={\faUser\ \textsc{Prompt for Qwen3-4B}},
  fonttitle=\bfseries\footnotesize,
  listing engine=listings,
  listing only,
  listing options={
    basicstyle=\ttfamily\scriptsize,
    breaklines=true,
    breakatwhitespace=false,
    columns=fullflexible,
    keepspaces=true,
    showstringspaces=false,
    aboveskip=0pt,
    belowskip=0pt,
    breakindent=0pt,
    postbreak=\mbox{},
    style=llmmono,
    literate={，}{,}1 {。}{.}1 {：}{:}1 {；}{;}1{（}{{(}}1 {）}{{)}}1 {【}{{[}}1 {】}{{]}}1 {、}{,}1,
  },
  colback=black!2, colframe=black!20, borderline={0.5mm}{0mm}{black!15},
  before skip=8pt, after skip=8pt,
  breakable,
  enhanced
}

\newtcolorbox{PureBox}{
  enhanced jigsaw,
  breakable,
  colback=white,
  colframe=black,
  boxsep=4pt, left=4pt, right=4pt, top=4pt, bottom=4pt,
  verbatim,              
}

\newtcblisting{tooldesc}{
  title={ \textsc{Tool Configuration}},
  fonttitle=\bfseries\footnotesize,
  listing engine=listings,
  listing only,
  listing options={
    basicstyle=\ttfamily\scriptsize,
    breaklines=true,
    breakatwhitespace=false,
    columns=fullflexible,
    keepspaces=true,
    showstringspaces=false,
    aboveskip=0pt,
    belowskip=0pt,
        breakindent=0pt,
    postbreak=\mbox{},   
  },
  colback=black!2, colframe=black!20, borderline={0.5mm}{0mm}{black!15},
  before skip=8pt, after skip=8pt,
  breakable,
  enhanced
}

\newtcblisting{hps}{
  title={ \textsc{Hyperparameters for RL}},
  fonttitle=\bfseries\footnotesize,
  listing engine=listings,
  listing only,
  listing options={
    basicstyle=\ttfamily\scriptsize,
    breaklines=true,
    breakatwhitespace=false,
    columns=fullflexible,
    keepspaces=true,
    showstringspaces=false,
    aboveskip=0pt,
    belowskip=0pt,
        breakindent=0pt,
    postbreak=\mbox{},   
  },
  colback=black!2, colframe=black!20, borderline={0.5mm}{0mm}{black!15},
  before skip=8pt, after skip=8pt,
  breakable,
  enhanced
}

\newtcblisting{promptours}{
  title={\faUser\ \textsc{Prompt for Ours}},
  fonttitle=\bfseries\footnotesize,
  listing engine=listings,
  listing only,
  listing options={
    basicstyle=\ttfamily\scriptsize,
    breaklines=true,
    breakatwhitespace=false,
    columns=fullflexible,
    keepspaces=true,
    showstringspaces=false,
    aboveskip=0pt,
    belowskip=0pt,
        breakindent=0pt,
    postbreak=\mbox{},   
  },
  colback=black!2, colframe=black!20, borderline={0.5mm}{0mm}{black!15},
  before skip=8pt, after skip=8pt,
  breakable,
  enhanced
}

\newtcblisting{respqwen}{
  title={\faTimes\ \textsc{Response From Qwen3-4B}},
  fonttitle=\bfseries\footnotesize,
  listing engine=listings,
  listing only,
  listing options={
    basicstyle=\ttfamily\scriptsize,
    breaklines=true,
    breakatwhitespace=false,
    columns=fullflexible,
    keepspaces=true,
    showstringspaces=false,
    aboveskip=0pt,
    belowskip=0pt,
        breakindent=0pt,
    postbreak=\mbox{},   
  },
  colback=black!2, colframe=black!20, borderline={0.5mm}{0mm}{black!15},
  before skip=8pt, after skip=8pt,
  breakable,
  enhanced
}

\newtcolorbox{StepBox}[1]{
  title={\textsc{#1}}, fonttitle=\bfseries\footnotesize,
  colback=black!2, colframe=black!20, borderline={0.5mm}{0mm}{black!15},
  before skip=8pt, after skip=8pt
}

\newtcblisting{LLMjudgeBox}{
  title={\ding{220}\; LLM-as-SQL-Error-Classifier}, colback=ThoughtBg, colframe=ThoughtBd, borderline={0.6mm}{0mm}{ThoughtBd},
  listing engine=listings, listing only,   listing options={style=llmmono},
}

\usepackage{soul}
\usepackage{tikz}
\usetikzlibrary{er,positioning,bayesnet}
\newcommand{\myfontsize}{\footnotesize} 
\definecolor{deepblue}{HTML}{1F4B99} 
\newcommand{\mytextbox}[2]{%
  \tikz[baseline=(X.base)]{%
    \node[
      draw=#1,
      thick,
      rounded corners,
      inner sep=2pt
    ] (X) {\myfontsize #2};%
  }%
}

\DeclareRobustCommand{\propose}{%
  \mytextbox{deepblue}{\texttt{PROPOSE}}\xspace%
}

\DeclareRobustCommand{\execute}{%
  \mytextbox{deepblue}{\texttt{EXECUTE}}\xspace%
}

\DeclareRobustCommand{\everify}{%
  \mytextbox{deepblue}{\texttt{E-VERIFY}}\xspace%
}

\DeclareRobustCommand{\mverify}{%
  \mytextbox{deepblue}{\texttt{M-VERIFY}}\xspace%
}

\DeclareRobustCommand{\selfcorrect}{%
  \mytextbox{deepblue}{\texttt{SELF-CORRECT}}\xspace%
}

\DeclareRobustCommand{\finalize}{%
  \mytextbox{deepblue}{\texttt{FINALIZE}}\xspace%
}

\definecolor{dustyblue}{HTML}{FCEF91}

\sethlcolor{dustyblue}
\usepackage{pifont}
\usepackage{graphicx} 

\usepackage{graphicx}   
\usepackage{hyperref}   
\usepackage{titling}    
\usepackage{threeparttable}
\definecolor{Qwen17}{RGB}{245,242,255} 
\definecolor{Qwen4B}{RGB}{237,250,242} 

\usepackage{fontawesome5}  
\usepackage{pifont}        
\hypersetup{colorlinks=true, linkcolor=blue, urlcolor=blue, citecolor=blue}

\usepackage{tikz}    
\usepackage{xcolor}  
\usepackage{hyperref}

\NewDocumentCommand{\circled}{O{0.6em} O{0.3ex} m}{%
  \tikz[baseline={([yshift=-#2]n.base)}]
    \node[
      draw,
      circle,
      minimum size=#1,
      inner sep=0.06ex,
      line width=0.4pt,
      font=\tiny\bfseries
    ] (n) {#3};%
}

\usepackage[nameinlink,noabbrev]{cleveref} 
\crefname{table}{Table}{Tables}   
\crefname{figure}{Fig.}{Figs.}    

\DeclareRobustCommand{\Finding}[3]{%
  \leavevmode
  \hypertarget{find:#1}{\textbf{\textcolor{teal}{Result~#1:}}}~#2%
  ~(\cref{#3})%
}

\DeclareRobustCommand{\FindingSimple}[3]{%
  \leavevmode
  \hypertarget{find:#1}{\textbf{\textcolor{teal}{Result~#1:}}}~#2%
  ~(\cref{#3})%
}

\newcommand{\githubc}[3][]{%
  \href{#2}{{\color{#3}\faGithub}\ #1}%
}

\usepackage{tablefootnote}

\usepackage{graphicx}
\newcommand{\TitleLogo}{%
  \raisebox{-2.2ex}{\includegraphics[height=3em]{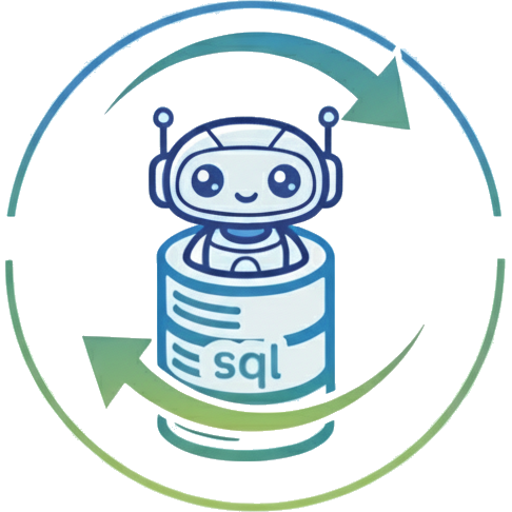}}%
}
\vspace{-0.1in}
\title{\TitleLogo\hspace{0.2em} MTSQL-R1: Towards Long-Horizon Multi-Turn Text-to-SQL via Agentic Training}

\author{
 \textbf{Taicheng Guo\textsuperscript{1,2}\thanks{Work done during internship at Amazon.}},
 \textbf{Hai Wang\textsuperscript{2}},
 \textbf{ChaoChun Liu\textsuperscript{2}},
  \textbf{Mohsen Golalikhani\textsuperscript{2}},  \\ 
 \textbf{Xin Chen\textsuperscript{3}\thanks{Work done at Amazon.}}, 
\textbf{Xiangliang Zhang\textsuperscript{1}},
  \textbf{Chandan K. Reddy\textsuperscript{2}}
\\[3pt] 
 \textsuperscript{1} University of Notre Dame,
 \textsuperscript{2} Amazon,
 \textsuperscript{3} Salesforce
\\ 
 \small{
   \textbf{Correspondence:} 
   tguo2@nd.edu, ckreddy@amazon.com, xzhang33@nd.edu
 } \\[3pt]
\githubc[https://github.com/taichengguo/MTSQL-R1]{https://github.com/taichengguo/MTSQL-R1}{black}\\
} 

\begin{document}
\maketitle


\begin{abstract}

Multi-turn Text-to-SQL aims to translate a user’s conversational utterances into executable SQL while preserving dialogue coherence and grounding to the target schema. However, most existing systems only regard this as a simple text translation task and follow a short-horizon paradigm, generating a query per turn without execution, explicit verification, and refinement, which leads to non-executable or incoherent outputs.
We present \textsc{MTSQL-R1}, an agentic training framework for \textit{long-horizon multi-turn Text-to-SQL}. We cast the task as a Markov Decision Process (MDP) in which an agent interacts with (i) a database for execution feedback and (ii) a persistent dialogue memory for coherence verification, performing an iterative \emph{propose$\to$execute$\to$verify$\to$refine} cycle until all checks pass.
Experiments on CoSQL and SParC demonstrate that \textsc{MTSQL-R1} consistently outperforms strong baselines, highlighting the importance of environment-driven verification and memory-guided refinement for conversational semantic parsing. Full recipes will be released upon acceptance to contribute to community research.
\end{abstract}

\input{latex/1-Introduction}
\input{latex/2-Related_Work}
\input{latex/3-methodology}

\input{latex/4-Experiments}
\input{latex/5-Conclusion}

\section*{Limitations}
While our method attains state-of-the-art performance with smaller model sizes, residual errors remain, notably Aggregation Drift (as shown in Fig.~\ref{fig:sql_error}), and some extra-hard cases (as shown in Fig.~\ref{fig:diff-pair}) are still unresolved. 
While our method achieves strong performance gains and lies on the Pareto frontier, it incurs higher latency and token usage compared to baselines. This may limit its applicability in real-time settings. We leave these challenges to future work toward more capable Text-to-SQL models.

\bibliography{custom}

\input{latex/Appendix}

\end{document}

%% file: latex/1-Introduction.tex
\section{Introduction}


\begin{table*}[t]
\vspace{-0in}
\centering
\setlength{\tabcolsep}{6pt}
\renewcommand{\arraystretch}{1.2}
\resizebox{\textwidth}{!}{%
\begin{tabular}{lcccc l}
\toprule
\textbf{Method} & \textbf{Conversation} & \textbf{Tool (DB) Integrated} & \textbf{Coherence Verification} & \textbf{Main Contributions} & \textbf{Base Model} \\
\midrule
Reasoning-SQL~\cite{pourreza2025reasoningsqlreinforcementlearningsql} & Single & ✗ & ✗ & RL (GRPO) & Open-source LLM \\
SQL-R1~\cite{ma2025sqlr1trainingnaturallanguage} & Single & ✗ & ✗ & RL (GRPO) & Open-source LLM \\
CoE-SQL~\cite{co2023coesql} & Multi & ✗ & ✗\,(Implicit Edit) & Edit-based Prompting & Closed-source (GPT-3.5/4) \\
ACT-SQL~\cite{zhang2023actsql} & Multi & ✗ & ✗ & Auto-CoT Prompting & Closed-source (GPT-3.5/4) \\
\rowcolor{Qwen4B} \textbf{\textsc{MTSQL-R1} (Ours)} & \textbf{Multi} & \textbf{✓} & \textbf{✓} & \textbf{Warm-Start SFT + Multi-Turn RL} & \textbf{Open-source LLM} \\
\bottomrule
\end{tabular}%
}\vspace{-0.1in}
\caption{\textbf{Comparison of  Text-to-SQL approaches.} 
\textsc{MTSQL-R1} integrates long-horizon formulation and enables multi-turn Text-to-SQL training, while all prior works rely on short-horizon and prompting/single-turn training. } 
\label{tab:method-compare}
\vspace{-0.1in}
\end{table*}

Multi-turn Text-to-SQL requires mapping each utterance to a SQL query while maintaining cross-turn coherence and schema grounding. Compared to single-turn settings, it demands robust handling of long-range dependencies under evolving user intents and previously issued constraints. Recent studies have explored the potential of LLMs for this task. Prompt-based LLM agents such as CoE-SQL~\cite{co2023coesql} and ACT-SQL~\cite{zhang2023actsql} rely on in-context learning to condition generation on dialogue history. Meanwhile, reasoning-oriented approaches such as Reasoning-SQL~\cite{pourreza2025reasoningsqlreinforcementlearningsql} and SQL-R1~\cite{ma2025sqlr1trainingnaturallanguage} show promise for single-turn text-to-SQL using reinforcement learning, yet still treat it purely as a translation task without interacting with the database environment. Although multi-turn Text-to-SQL has attracted increasing attention, existing methods share a critical limitation: they operate under a \emph{short-horizon reasoning} paradigm.

\begin{figure}[t]
\centering
\includegraphics[width=0.47\textwidth]{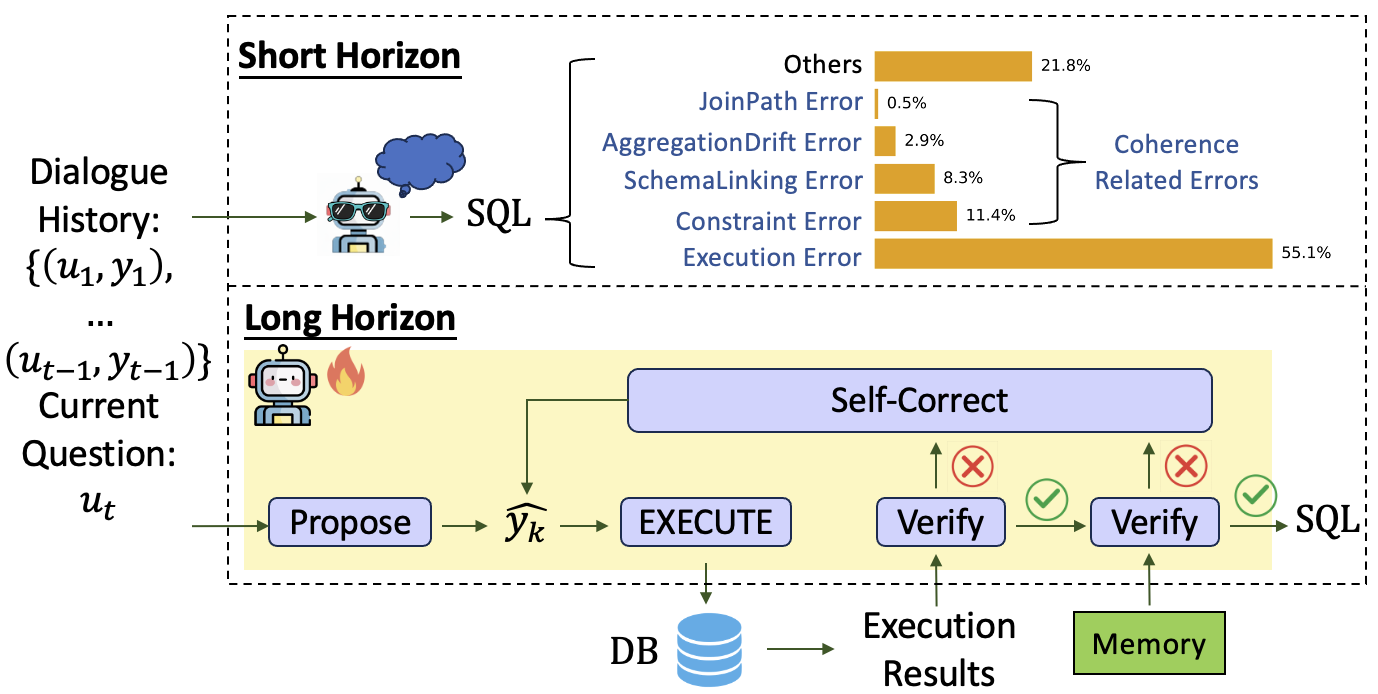}\vspace{-0in}
\caption{\textbf{Short- vs. long-horizon modeling in multi-turn Text-to-SQL.}
Short-horizon models directly translate text to SQL (causing a large portion of execution error), while our long-horizon  \textsc{MTSQL-R1} interacts with the database and the maintained dialogue memory for executable and consistent queries.}
\vspace{-0.1in}
\label{fig:intro_problem} 
\end{figure}

Short-horizon reasoning generates SQL queries using only the current utterance and minimal prior context (see Fig.~\ref{fig:intro_problem}). This limitation manifests in two ways: 
(1) \textit{Lack of verification:} Models never interact with the database for explicit verification and perform explicit checks for historical dialogue and schema coherence, leading to semantically invalid or inconsistent outputs. 
(2) \textit{Lack of correction:} Without explicit and detailed verification feedback, models struggle to iteratively correct earlier incorrect SQL queries.

To address these issues, we introduce \textsc{MTSQL-R1}, an agentic training framework for long-horizon multi-turn Text-to-SQL. By long-horizon reasoning, we mean explicitly verifying intermediate predictions through environment interactions and performing self-correction based on the resulting signals. Specifically, our approach enables:
\begin{itemize}[leftmargin=*]
    \item 
\textbf{Environment-based verification:} The model interacts dynamically with two components: (i) a database for execution feedback and (ii) a long-term dialogue memory for explicit coherence checking to verify intermediate SQL outputs.
    \item \textbf{Self-correction:} Based on verification feedback, the model iteratively refines its generated SQL queries to achieve consistent, executable outputs across multiple turns.
\end{itemize}

To realize this capability, \textsc{MTSQL-R1} is built in three stages: \textbf{1) Problem formulation:} We define multi-turn Text-to-SQL as a Markov Decision Process (MDP) with environment-driven feedback. \textbf{2) Warm-Start supervised fine-tuning (SFT):} We synthesize high-quality long-horizon trajectories via a self-taught exploration procedure with rejection sampling and use them to initialize the model. \textbf{3) End-to-end reinforcement learning (RL):} The SFT model is further optimized with multi-level rewards derived from execution success and memory coherence, enhancing its ability to verify and self-correct autonomously.
We evaluate \textsc{MTSQL-R1} on CoSQL and SParC benchmarks. Using 1.7B- and 4B-parameter backbones, our models achieve state-of-the-art results. Our key contributions are:
\begin{itemize}[leftmargin=*]
    \item We propose \textsc{MTSQL-R1}, the first multi-turn Text-to-SQL framework with explicit execution- and memory-based verification and self-correction (Table~\ref{tab:method-compare}).
    \item We introduce a long-horizon training pipeline combining self-taught Warm-Start SFT with end-to-end RL using multi-level rewards.
    \item We demonstrate consistent gains in coherence, executability, and generalization, revealing fresh insights into long-horizon multi-turn Text-to-SQL.
\end{itemize}

%% file: latex/2-Related_Work.tex
\section{Related Work}
\vspace{-0in}
\paragraph{Multi-turn Text-to-SQL:}
Methods for multi-turn text-to-SQL can be divided into pre-LLM and LLM-based methods. Pre-LLM approaches focused on specialized neural architectures for modeling dialogue and schema context, leveraging prior SQL~\cite{zhang2019editing, wang2020rat}, graph-based representations~\cite{cai2020igsql}, or dynamic schema-linking~\cite{hui2021r2sql, zheng2022hie}. RASAT~\cite{qi2022rasat} enhanced Transformers with relation-aware attention and syntactic constraints~\cite{scholak2021picard}.
LLM-based methods instead rely on prompting: ACT-SQL~\cite{zhang2023actsql} rewrites multi-turn queries into single-turn inputs via chain-of-thought prompting, while CoE-SQL~\cite{co2023coesql} edits prior SQL incrementally. Both depend on closed-source GPT models and lack database verification or self-correction.

\paragraph{Reasoning Models for Single-Turn Text-to-SQL:}

Recent reasoning-oriented models target single-turn Text-to-SQL. STaR-SQL~\cite{he2025starsqlselftaughtreasonertexttosql} uses rationale-based SFT, while Reasoning-SQL~\cite{pourreza2025reasoningsqlreinforcementlearningsql} and SQL-R1~\cite{ma2025sqlr1trainingnaturallanguage} apply reinforcement learning for logical and execution consistency. However, they omit dialogue coherence and interactive verification, making them unsuitable for multi-turn reasoning.

\paragraph{Long-Horizon Reasoning with RL:}

RL has advanced long-horizon reasoning in LLMs such as OpenAI’s O-series~\cite{openai_o_series_2025}, DeepSeek-R1~\cite{guo2025_deepseekR1}, and Kimi K1.5~\cite{kimi2025_k1_5}. Models like Search-R1~\cite{jin2025_searchR1} and WebAgent-R1~\cite{wei2025_webagentR1} extend reasoning via environment interaction. Yet, none operates in the context of multi-turn Text-to-SQL. 

%% file: latex/3-methodology.tex
\section{Methodology}
\vspace{-0in}

\begin{figure*}[t]
\vspace{-0in}
\centering
\includegraphics[width=0.88\textwidth]{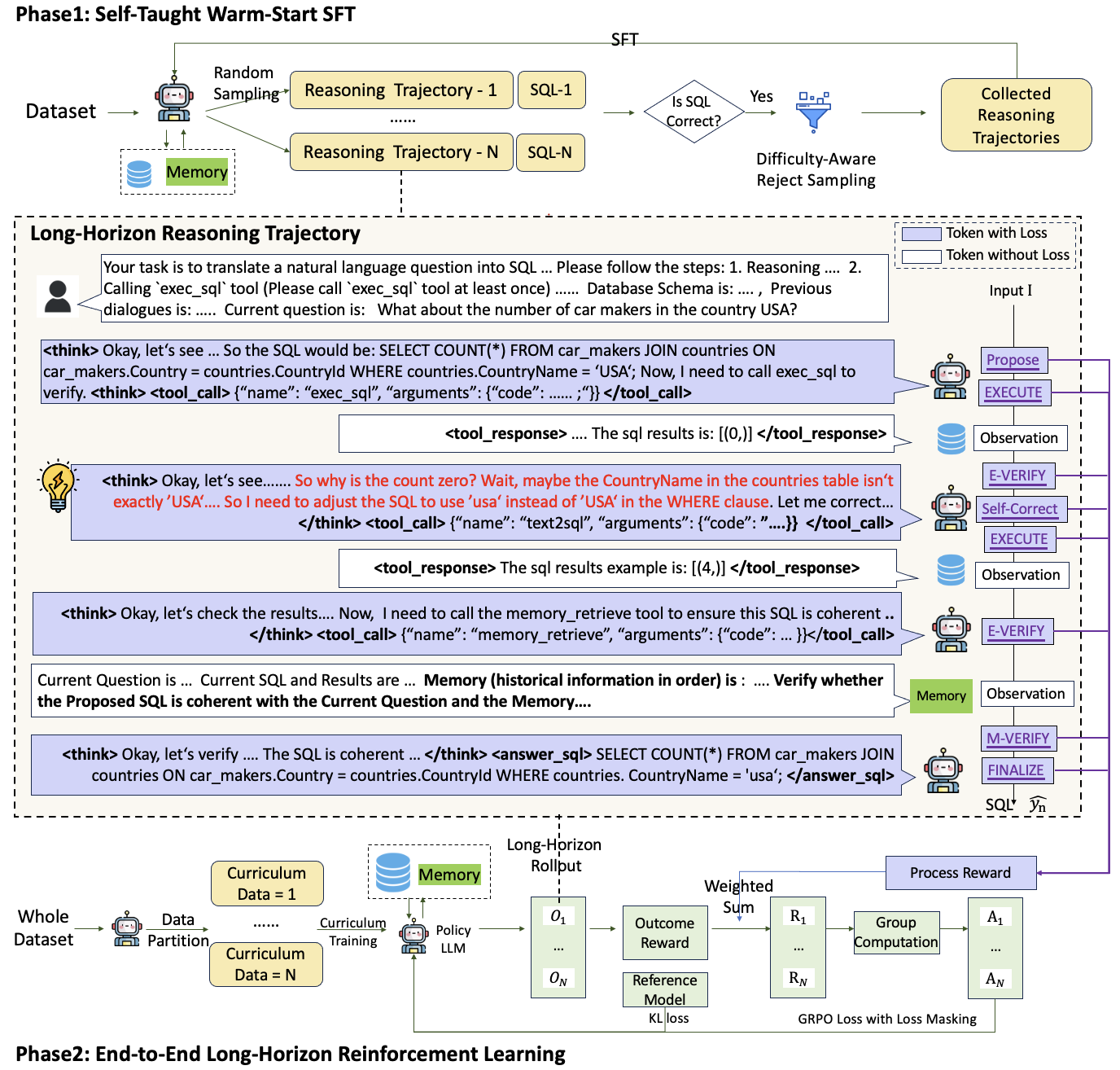} \vspace{-0in}
\caption{\textbf{Overview of the \textsc{MTSQL-R1} training pipeline.} (1) Phase 1 (Self-Taught warm-start SFT): verified multi-turn trajectories are generated to provide initial supervision for warm-start fine-tuning. (2) \textbf{Aha-moment trajectory}: an illustrative long-horizon Text-to-SQL example generated by the final RL-trained model, shown to clarify the trajectory format. (3) Phase 2 (End-to-End long-horizon RL): the policy LLM interacts with the database and memory over multiple turns and is optimized with multi-turn RL to strengthen long-horizon reasoning.}
\vspace{-0.2in}
\label{fig:SFT} 
\end{figure*}  


\subsection{Problem Formulation}
Let the dialogue up to turn $t-1$ be denoted as $H_{t-1}=\{(u_1,y_1),\ldots,(u_{t-1},y_{t-1})\}$, where $u_i$ is the user utterance and $y_i$ is the SQL at turn $i$.
The goal of multi-turn Text-to-SQL is: given $H_{t-1}$ and the current utterance $u_t$, generate the SQL $y_t$ for turn $t$.
Prior work commonly treats the task as direct translation with a policy $\pi_\theta$:
$\{H_{t-1},u_t\}\xrightarrow{\ \pi_\theta\ } y$, 
without modeling intermediate reasoning or long-term planning. Such short-horizon solutions ignore iterative verification and self-correction signals that are crucial for complex, multi-turn scenarios.

\paragraph{Our Long-Horizon Formulation:}  
We cast multi-turn Text-to-SQL as a Markov Decision Process (MDP) with policy $\pi_\theta$:

\begin{itemize}[leftmargin=*]
    \item \textbf{Environment:} We set up two environment components: (i) A relational database $D=(S,T)$ (schema $S$ and tables $T$) for SQL execution; (ii) a maintained \textbf{long-term dialogue memory} $M_{t-1}$ that stores, up to turn $t$, questions $u_i$, SQL $y_i$, and tool-parsed constraints/entities $m_i$ for later self-verification.
    \item \textbf{Inner step ($k$).} An inner reasoning step.
    \item \textbf{State.} $s_k=(H_{t-1},S,u_t,M_{t-1},\hat y_k,\mathrm{obs}_{1:k-1})$, where $\hat y_k$ is the intermediate SQL and $\mathrm{obs}_{1:k-1}$ are accumulated execution results/errors.
    \item \textbf{Action space $a_k\in\mathcal A$.}
    \begin{enumerate}[leftmargin=*]
        \item \propose:  directly attempt to generate SQL $\hat{y}_k$ given the initial state $s_0$;
        \item \execute: run $\hat y_k$ on $D$ to obtain resulting rows or error messages;
        \item \everify: judge execution-based correctness after \execute;
         \item \mverify: check $\hat y_k$ against $M_{t-1}$ for cross-turn coherence (constraints/entities);
        \item \selfcorrect: refine $\hat y_k$;
        \item \finalize: commit $\hat y_k$ as $y$ and terminate the episode.
    \end{enumerate}  
    \item \textbf{Observation.} Determined by the preceding action (e.g., \execute yields results/errors; at the start of \mverify we compute a violation set).
    \item \textbf{Transition ($\mathcal{P}(s_{k+1} \mid s_k, y_k)$): } Deterministic for non-execution actions; environment-driven for \execute.
    \item \textbf{Policy} $\pi_\theta(a_k\mid s_k)$ over discrete actions; the LLM generates textual content for \propose, \everify, \mverify, and \selfcorrect. The policy is autonomously learned by the following training recipes, including Warm-Start SFT and end-to-end RL.
    \item \textbf{Objective:} Maximize expected reward, measuring the correctness of the final SQL. This MDP formulation enables iterative \emph{propose$\to$execute$\to$verify$\to$refine} cycles until all checks pass.
\end{itemize}  
\vspace{-0em}
Concretely, as shown in~\Cref{fig:SFT}, $\hat{y}_k$ is an
intermediate SQL query and $y$ is the final executable SQL. Either verification
can loop back to $\hat{y}_k$, yielding iterative refinement until all checks pass.

\subsection{Warm-Start SFT for Behavior Cloning}
\label{sec:warmstart}


\subsubsection{Data Formats}

To incorporate long-horizon reasoning patterns into LLM, we first propose the following Long-Horizon SFT dataset format and the loss masking technique to achieve the \textit{behavior cloning} for the agent.
We construct SFT trajectories that strictly follow the MDP (Fig.~\ref{fig:SFT}), capturing the full episode $\big(I,a_1,\hat y_1,\mathrm{obs}_1,\ldots,a_n,\hat y_n\big)$
where $I$ is the packed instruction/prompt. The input includes: (1) system instructions; (2) the current question $u_t$, dialogue $H_{t-1}$, and schema $S$; (3) tool instructions: \execute and \mverify are treated as tool functional calling to the environment (database and memory, respectively). The action transition rule is:

{\tiny
\setlength{\abovedisplayskip}{2pt}
\setlength{\belowdisplayskip}{2pt}
\setlength{\abovedisplayshortskip}{1pt}
\setlength{\belowdisplayshortskip}{1pt}
\begin{equation}
\text{Type}(a_{k+1}) =
\begin{cases}
\text{PROPOSE}, \quad  a_k=\varnothing \ \text{(Initial state)},\\
\text{EXECUTE} , \quad  \text{Type}(a_k)\in\{\text{PROPOSE},\text{SELF-CORRECT}\},\\
\text{E-VERIFY} , \quad  \text{Type}(a_k)=\text{EXECUTE},\\
\text{M-VERIFY} , \quad  \text{Type}(a_k)=\text{E-VERIFY and} \; \hat{y}_k \; \text{passes},\\
\text{SELF-CORRECT}, \ \text{Type}(a_k)\in\{\text{E/M-VERIFY}\}\ \text{and} \; \hat{y}_k \; \text {fails},\\
\text{FINALIZE}, \quad  \text{Type}(a_k)=\text{M-VERIFY and}  \; \hat{y}_k \; \text{passes}.
\end{cases}
\label{eq:transition}
\end{equation}
}

Following the transition rule in~\Cref{eq:transition}, given input $I$, the language agent will first \propose  an initial SQL $\hat{y}_k$, then \execute  it against the database to obtain execution feedback $\mathrm{obs}_k$. It next performs \everify to assess correctness from the feedback and \mverify to check consistency between $\hat{y}_k$ and the long-term memory $M_{t-1}$, ensuring logical coherence and avoiding contradictions. If $\hat{y}_k$ fails either verification, the agent enters \selfcorrect to refine $\hat{y}_k$ and repeats the verify-correct loop. The long-horizon SFT dataset is collected autonomously from the agent’s MDP rollouts and represented as a text trajectory. 

\paragraph{Loss Masking.}
To teach behaviors rather than memorize observations, we mask tokens from instructions $I$, execution outputs $\mathrm{obs}$, and memory prompts, supervising only actions and SQL:
$\mathcal{L}_{\text{SFT}} = -\sum_{t=1}^T m_t \log \pi_\theta(w_t\mid w_{<t},I)$, where $m_t=1$ if $w_t$ is an action or SQL token.



\subsubsection{Self-Taught Warm-Start SFT}

\paragraph{Single-Round Trajectory Collection.}

With a long-horizon MDP setup, we first prompt the base LLM on all training questions to generate trajectories, retain only those that yield correct SQL as target behaviors, and fine-tune on them to initialize.


\paragraph{Why Self-Taught?}
Even with multiple samples per question, the base model leaves many cases unsolved, limiting coverage of high-quality trajectories. Simply pairing a question with the gold SQL to synthesize a trajectory fails to reflect natural execution errors. We therefore introduce a \emph{self-taught} iterative procedure that continually strengthens the model and expands the pool of verified trajectories. Let \(i\) index the iteration and \(\pi_{\theta_i}\) be the model used both to generate trajectories and to undergo fine-tuning. We maintain: (i) \(D_i\), the training subset used to synthesize trajectories, and (ii) $\Traj$, the cumulative set of trajectories for fine-tuning. The overall process is shown in Algorithm~\ref{ws_sft}. 

The algorithm consists of four stages. \textbf{S1 Trajectory Collection: } For each training instruction, generate 20 rollouts from the current policy at temperature 0.7 and keep only those whose final SQL matches the gold query. \textbf{S2 Difficulty-Aware Rejection Sampling: } Among trajectories whose final SQL is correct, we perform difficulty-aware rejection sampling. The intuition is that not every query requires long-horizon reasoning: we want long and diverse trajectories for hard cases, and short, deterministic ones for simple cases. We determine difficulty using (i) standard SQL hardness criteria (e.g., Spider) and (ii) the current model’s competence. For items that are \emph{easy} or perfectly solved across 20 samples, we randomly keep up to two short trajectories ($\leq$2 interactions). For \emph{hard} items, we retain longer trajectories ($\ge$2 interactions), cluster them with Qwen3-Embedding~\cite{yang2025qwen3technicalreport}, and sample three representatives. \textbf{S3 SFT} to update $\pi_\theta$; 
\textbf{S4 Dataset Update:} The training dataset is updated by removing all instructions that already produced high-quality trajectories in the current round, yielding $D_{i+1}$. We repeat the process until reaching the maximum number of rounds. 


\subsection{Long-Horizon End-to-End RL}
\label{sec:rl}
\subsubsection{Curriculum RL Training}
In LLM RL training, Extra-hard SQL queries induce \emph{overly sparse rewards} and long-horizon credit-assignment challenges, making exploration unstable for policy optimization. A curriculum mitigates this by scheduling training from easier to harder instances, which is known to yield faster and more reliable convergence. We therefore adopt an easy\(\rightarrow\)hard curriculum for RL training.
For each training example, we sample 20 trajectories and compute a success count: 
$s_i = \#\{\text{correct out of 20, measured by EX \emph{and} EM}\}$.
We discard examples with $s_i = 20$ (too easy). The remaining examples are sorted in descending order by $s_i$ (higher = easier) and partitioned into contiguous bins of size 2000. We label the bins as curriculum levels, with \textit{Curriculum Data = 1} denoting the easiest set.
During RL, the policy $\pi_\theta$ interacts with tools following the MDP loop to produce trajectories. Database and memory interactions supply grounded signals that drive verification and self-correction.

\subsubsection{Reward}

\paragraph{Why do we need Multi-level rewards?}
In the long-horizon MDP, the agent generates a trajectory.
A terminal reward on $\hat{y}_n$ is too sparse, especially for hard cases, making them hard to learn from.
We therefore introduce multi-level rewards with outcome and dense process-level feedback, guiding stepwise reasoning rather than only the final answer.
We first present the rule-based outcome reward, then the process reward.

\paragraph{Execution Match (EX) Reward and Exact Match (EM) Reward.}
To align the agent’s SQL with the user intent, we execute the prediction $\hat{y}_n$ and compare its result with the ground-truth $y$: 
$\mathcal{R}_{\text{EX}}(\hat{y_n}, y) = \mathcal{I} (\texttt{Exec}(\hat{y}_n) == \texttt{Exec}(y))$.
Matching outputs yield reward 1; otherwise 0. 
Here, \texttt{Exec}(SQL) denotes the query’s execution result on the database, and $\mathcal{I}$ is the indicator function. We also use a strict string-level signal that requires the predicted SQL to exactly match the reference (including order, formatting, etc):
$\mathcal{R}_{\text{EM}}(\hat{y_n}, y) = \mathcal{I} (\hat{y}_n == y)$.

\paragraph{Process Reward Design Principle.}
Because the agent autonomously generates trajectories
$(I, a_1, \hat{y}_1, \mathrm{obs}_1, a_2, \dots, a_n, \hat{y}_n)$, our process reward supervises how each action type, including \propose, \everify, \mverify and \selfcorrect, should behave based on the quality of its immediate outcome. In other words, relative to the previous step, does this step move the solution closer to the goal? Accordingly, we treat each action $a$ in the trajectory as a sub-process and define an action-level reward function specific to its type:

\begin{itemize}[leftmargin=*]
  \item \propose and \selfcorrect: For these actions, the process result is the candidate SQL $\hat{y}_k$. Hence, we design Clause Match as a dense reward to measure how well the predicted query aligns with the gold query across major SQL clauses: $\mathcal{R}(a_k\mid\hat{y})_{\text{Propose/Self-Correct}} =
\operatorname{AVG}\,\mathrm{F1}\!\left(c(\hat{y}_k),\, c(y)\right)$, where \(c\) ranges over the SQL clauses
\texttt{SELECT}, \texttt{WHERE}, \texttt{JOIN}, \texttt{GROUP}, \texttt{ORDER}. $F1$ is the F1-Score calculation.

  \item \everify and \mverify: For these actions, the process result is whether the verification is correct.
  We require the model to output a binary flag $\texttt{VR}\in\{\texttt{pass},\texttt{fail}\}$ that states the verdict.
  Let $\hat{y}_{k-1}$ be the SQL being verified. For \everify, we have: $\mathcal{R}\!\left(a_k \,\middle|\, \hat{y}_{k-1}\right)_{\text{E-Verify}}=\text{the entry at }(\text{Exec Results, VR})$ in:
\vspace{-0in}
\begin{equation}
\notag
\label{eq:exec-reward-lookup}
\def\ok{\mathrm{ok}}\def\null{\mathrm{null}}\def\err{\mathrm{error}}
\renewcommand{\arraystretch}{1.05} 
\scriptsize                        
\begin{array}{c} 
\begin{array}{c|cc}
 & VR=fail & VR=pass\\ \hline
\text{Exec Results}=\ok    & 0   & 1 \\
\text{Exec Results}=\null  & 0.1 & 0 \\
\text{Exec Results}=\err   & 1   & 0 \\
\end{array}
\\ 
\end{array}
\end{equation}
  
  For \mverify, we have: 
  {\scriptsize
  \begin{equation}
  \mathcal{R}(a_k \mid \hat{y}_{k-1})_{\text{M-Verify}} =
  \begin{cases}
   \operatorname{AVG}\,\mathrm{F1}\!\left(c(\hat{y}_k),\, c(y)\right),  \ \text{if } \texttt{VR}=\texttt{pass} \\[3pt]
  1 - \operatorname{AVG}\,\mathrm{F1}\!\left(c(\hat{y}_k),\, c(y)\right), \ \text{otherwise},
  \end{cases}
  \nonumber
  \end{equation} 
  }\normalsize
where \(c\) ranges over the SQL clauses \texttt{SELECT}, \texttt{WHERE}, \texttt{JOIN}, \texttt{GROUP}, and \texttt{ORDER}, and \(\hat{y}_k\) denotes the candidate SQL evaluated in this verification. Intuitively, a higher reward indicates that \(\hat{y}_k\) is more consistent with the verification outcome.
\end{itemize}

Finally, for simplicity, given a whole trajectory, we take a \emph{weighted sum} of all outcome-level and process-level rewards defined above. The weights are selected via grid search on a small held-out subset of the training data (used as a validation set).
$\mathcal{R}_{all} = w_1 * (\mathcal{R}_{\text{EX}} + \mathcal{R}_{\text{EM}}) + w_2 * (\mathcal{R}_{\text{Propose/Self-Correct}} + \mathcal{R}_{\text{E-Verify}} + \mathcal{R}_{\text{M-Verify}})$.

  




\subsubsection{GRPO Training with Loss Masking}

Following~\cite{shao2024deepseekmathpushinglimitsmathematical}, for each question we sample \(G\) trajectories \(\{O_i\}_{i=1}^G\), where \(O_i=(I,a_1,\hat{y}_1,\mathrm{obs}_1,\ldots,a_n,\hat{y}_n)\). Each trajectory receives a scalar reward \(r_i\); letting \(\mathbf{r}=(r_1,\ldots,r_G)\), we compute a group-normalized advantage shared by all tokens of trajectory \(i\):
$A_{i,t} = \frac{r_i - \operatorname{mean}(\mathbf{r})}{\operatorname{std}(\mathbf{r}) + \varepsilon}, \, \forall t$.
Thus, every token in a trajectory uses its normalized reward as the advantage.
Given the above advantages, we apply loss masking to exclude SQL execution outputs and human instruction tokens from the loss, so the model focuses on learning the reasoning process. The optimized GRPO loss is:
\begin{equation}
{\tiny
\begin{aligned}
\mathcal{J}_{\text{GRPO}}(\theta) 
&= \mathbb{E} \Bigg[ 
\frac{1}{G} \sum_{i=1}^{G} \frac{1}{|\mathcal{M}_i|} \sum_{t \in \mathcal{M}_i}
\Bigg\{ \min \Big[
r_{i,t} A_{i,t}, \\
&\hspace{-3em} \text{clip} \left(
r_{i,t},
1 - \epsilon,\ 1 + \epsilon
\right) A_{i,t}
\Big] \Bigg\} 
- \beta \mathbb{D}_{\text{KL}} \left[ \pi_{\theta} \parallel \pi_{\text{ref}} \right]
\Bigg] 
\end{aligned}
}
\nonumber
\end{equation}
where $G$ is the number of sampled trajectories per group; $r_{i,t}$=$\frac{\pi_{\theta}(o_{i,t}\mid q,o_{i,<t})}{\pi_{\theta_{\mathrm{old}}}(o_{i,t}\mid q,o_{i,<t})}$ is the per-token importance ratio;  $A_{i,t}$ is the token-level advantage. Following standard GRPO, we also apply a token mask $\mathcal{M}_i$ (keep only reasoning tokens).

%% file: latex/4-Experiments.tex
\section{Experiments}

\begin{table*}[t]
\vspace{-0in}
\centering
\small
\setlength{\tabcolsep}{5.2pt}
\renewcommand{\arraystretch}{1.15}
\begin{threeparttable}
\resizebox{0.96\textwidth}{!}{%
\begin{tabular}{llcccccccccc}
\toprule
& & \multicolumn{4}{c}{In-domain (\%)} & \multicolumn{4}{c}{Out-of-domain (\%)} & & \\
\cmidrule(lr){3-6}\cmidrule(lr){7-10}
\multirow{2}{*}{Model} & \multirow{2}{*}{Model Size} & \multicolumn{2}{c}{CoSQL} & \multicolumn{2}{c}{SParC} & \multicolumn{2}{c}{CoSQL} & \multicolumn{2}{c}{SParC} & Avg EX$\uparrow$ & Avg EM$\uparrow$ \\
& & EX$\uparrow$ & EM$\uparrow$ & EX$\uparrow$ & EM$\uparrow$ & EX$\uparrow$ & EM$\uparrow$ & EX$\uparrow$ & EM$\uparrow$ & & \\
\midrule
\multicolumn{12}{c}{\textit{Previous Reported Results (Frontier LLMs, CoT Prompting LLM Baselines, and Pre-LLM Baselines}} \\
GPT-4.1 & Closed-Source & 60.9 & 32.1 & 61.8 & 33.3 & — & — & — & — & 61.4 & 32.7 \\
OpenAI-O3 & Closed-Source & 59.8 & 29.1 & 57.0 & 30.3 & — & — & — & — & 58.4 & 29.7 \\
DeepSeek-R1 & 671B & 58.5 & 36.0 & 57.6 & 37.2 & — & — & — & — & 58.1 & 36.6 \\
Qwen-3-1.7B & 1.7B & 59.9 & 49.3 & 61.5 & 46.5 & — & — & — & — & 60.7 & 47.9 \\
Qwen-3-4B & 4B & 64.0 & 50.7 & 62.9 & 49.8 & — & — & — & — & 63.5 & 50.3 \\
Qwen-3-14B & 14B & 66.5 & 54.3 & 64.1 & 51.9 & — & — & — & — & 65.3 & 53.1 \\
Qwen-3-32B & 32B & 66.8 & 54.4 & 74.0 & 53.4 & — & — & — & — & 70.4 & 53.9 \\
\midrule
ACT-SQL~\cite{zhang2023actsql} & Closed-Source & 63.7 & 46.0 & 63.8 & 51.0 & — & — & — & — & 63.8 & 48.5 \\
CoE-SQL~\cite{co2023coesql} (Few-shot, 16-shot) & Closed-Source & 69.6 & 52.4 & 70.3 & 56.0 & 58.5 & 49.6 & 57.9 & 48.5 & 64.1 & 51.6 \\
Planning/Tool-using Agent built by LangGraph\tablefootnote{\url{https://docs.langchain.com/oss/python/langgraph/sql-agent}} & Closed-Source & 69.9 & 32.5 & 69.6 & 34.6 & — & — & — & — & 69.8 & 33.6 \\
\midrule
GAZP+BERT~\cite{zhong2020gazp} & $\sim$215M & 38.8 & 42.0 & 47.8 & 48.9 & — & — & — & — & 43.3 & 45.5 \\
HIE-SQL+GraPP~\cite{zheng2022hie} & $\sim$125M & — & 56.4 & — & 64.7 & — & — & — & — & — & 60.6 \\
RASAT+PICARD~\cite{qi2022rasat} & 3B & 67.0 & 58.8 & 73.3 & 67.7 & 55.8 & 48.0 & 61.9 & 56.1 & 64.5 & 57.7 \\
\midrule
\multicolumn{12}{c}{\textit{Short-Horizon Baselines and Our Long-Horizon Framework on Qwen-3-1.7B}} \\
\textbf{[Short-Horizon]} SFT & 1.7B & 68.1 & \underline{59.3} & \underline{74.3} & \textbf{69.2} & 64.1 & 55.2 & \underline{71.7} & \underline{65.1} & 69.6 & \underline{62.2} \\
\textbf{[Short-Horizon]} Direct RL (~\cite{pourreza2025reasoningsqlreinforcementlearningsql, ma2025sqlr1trainingnaturallanguage} & 1.7B & \underline{72.8} & 59.0 & 72.1 & 65.5 & \underline{66.3} & \underline{55.9} & 70.6 & 62.3 & \underline{70.5} & 60.7 \\
\rowcolor{Qwen17} \textbf{[Long-Horizon]} Base Agent without Training & 1.7B & 22.6 & 16.3 & 23.9 & 17.8 & — & — & — & — & 23.3 & 17.1 \\
\rowcolor{Qwen17} \textbf{[Long-Horizon]} Warm-Start SFT Only (Round 1) & 1.7B & 69.9 & 57.6 & 70.6 & 62.0 & 67.1 & 55.5 & 67.3 & 58.1 & 68.7 & 58.3 \\
\rowcolor{Qwen17} \quad + Warm-Start SFT Only (Round 2) & 1.7B & 72.2 & 60.5 & 72.3 & 63.0 & 67.2 & 54.2 & 70.0 & 61.5 & 70.4 & 59.8 \\
\rowcolor{Qwen17} \quad + Warm-Start SFT Only (Round 3) & 1.7B & 73.0 & 62.1 & 72.8 & 65.7 & 68.8 & 56.2 & 71.3 & 62.7 & 71.5 & 61.7 \\
\rowcolor{Qwen17} \textbf{[Long-Horizon]} Warm-Start SFT + RL (Outcome Only) & 1.7B & 76.6 & 62.7 & 76.2 & 66.1 & 70.3 & 59.8 & 73.0 & 66.2 & 74.0 \textcolor{red}{(+3.5)} & 63.7 \\
\rowcolor{Qwen17} \textbf{[Long-Horizon]} Warm-Start SFT + RL (Outcome + Process) & 1.7B & \textbf{77.3} & \textbf{63.5} & \textbf{76.2} & 66.1 & \textbf{70.4} & \textbf{59.8} & \textbf{74.5} & \textbf{68.0} & \textbf{74.6 \textcolor{red}{(+4.1)}} & \textbf{64.4} \\
\midrule
\multicolumn{12}{c}{\textit{Short-Horizon Baselines and Our Long-Horizon Framework on Qwen-3-4B}} \\
\textbf{[Short-Horizon]} SFT
& 4B
& 73.1 & \underline{64.8}
& \underline{78.3} & \textbf{71.5}
& 70.2 & 61.0
& \underline{75.1} & \underline{68.9}
& \underline{74.1} & \underline{66.6}
\\
\textbf{[Short-Horizon]} Direct RL (~\cite{pourreza2025reasoningsqlreinforcementlearningsql, ma2025sqlr1trainingnaturallanguage} & 4B & \underline{75.2} & \underline{64.8} & 75.8 & 66.5 & \underline{71.4} & \underline{61.3} & 73.4 & 64.0 & 74.0 & 64.2 \\
\rowcolor{Qwen4B} \textbf{[Long-Horizon]} Base Agent without Training & 4B & 60.3 & 45.6 & 57.6 & 44.1 & — & — & — & — & 59.0 & 44.9 \\
\rowcolor{Qwen4B}
\textbf{[Long-Horizon]} Warm-Start SFT Only (Round 1)
& 4B
& 73.9 & 62.1
& 73.8 & 63.1
& 72.7 & 58.7
& 74.0 & 64.0
& 73.6 & 62.0
\\
\rowcolor{Qwen4B}
\quad + Warm-Start SFT Only (Round 2)
& 4B
& 74.7 & 62.8
& 74.9 & 64.8
& 73.5 & 61.2
& 73.7 & 62.4
& 74.2 & 62.8
\\
\rowcolor{Qwen4B}
\quad + Warm-Start SFT Only (Round 3)
& 4B
& 75.2 & 63.0
& 75.1 & 65.6
& 72.3 & 61.8
& 74.0 & 64.4
& 74.2 & 63.7
\\
\rowcolor{Qwen4B}
\textbf{[Long-Horizon]} Warm-Start SFT + RL (Outcome Only)
& 4B
& 79.1 & 64.5
& 78.1 & 67.8
& 74.0 & 63.0
& 76.0 & 69.0
& 76.9 \textcolor{red}{(+2.8)} & 66.1
\\
\rowcolor{Qwen4B}
\textbf{[Long-Horizon]} Warm-Start SFT + RL (Outcome + Process)
& 4B
& \textbf{79.9} & \textbf{65.2}
& \textbf{79.0} & 68.7
& \textbf{74.0} & \textbf{63.0}
& \textbf{77.4} & \textbf{69.1}
& \textbf{77.6 \textcolor{red}{(+3.5)}} & \textbf{66.5} \\
\bottomrule
\end{tabular}
}
\end{threeparttable}
\caption{\textbf{Performance of our method.} In-Domain is the standard setting. The Out-domain (trained on one dataset and evaluated on another dataset) is designed to evaluate the generalization capability of different methods.  "—" denotes that the performance of Out-Domain is the same as In-Domain for methods that are not involved in training. \textcolor{red}{(+X.X)} denotes the improvement over the previous best baseline.}
\label{overall_perf}
\vspace{-0.2in}
\end{table*}


Our evaluation is organized into:

\begin{itemize}[leftmargin=*]
    \item \textbf{RQ1: Effectiveness, Generalization and Efficiency} 
    Does our long-horizon reasoning agent improve performance?  

    \item \textbf{RQ2: Evolution of Long-Horizon Capabilities.}  
    How do the agent’s long-horizon reasoning capabilities evolve during different training stages?  

    \item \textbf{RQ3: SQL Generation Quality.}  
    To what extent does the agent correctly or incorrectly predict different SQL syntactic structures, and what error patterns are reduced by our method?  

    \item \textbf{RQ4: Training Dynamics.} How stable is the training process? (Please See Appendix~\ref{sec:RQ4TD}) 
\end{itemize}

\subsection{Datasets, Implementation and Baselines}
We evaluate on two standard Text-to-SQL benchmarks: SParC~\cite{yu-etal-2019-sparc} and CoSQL~\cite{yu-etal-2019-cosql}. SParC includes 4,298 coherent question sequences (12,000+ questions) with paired SQL; CoSQL has 3,000 multi-turn dialogues with 10,000 annotated SQL. We report Execution Accuracy (EX) and Exact Match (EM), using the same definitions as in our reward design; implementation details appear in the Appendix~\ref{sec:Implementation_details}.

\paragraph{Baselines.}
\textbf{1) Frontier LLMs and reasoning models} include frontier LLMs such as GPT-4.1, and OpenAI-O3; \textbf{2) COT Prompting and RAG-Based LLM Baselines} include CoE-SQL \cite{co2023coesql}, which refines SQL queries across turns via chain-of-editing RAG prompting. ACT-SQL \cite{zhang2023actsql}, which generates chain-of-thoughts to guide complex reasoning, and  Planning/Tool-using Agent built with LangGraph on top of a strong closed-source model; \textbf{3) LLM Long-Horizon without Training} includes prompting non-fine-tuned reasoning base models to use the database and self-verification to verify the effectiveness of our training methods;  \textbf{4) LLM Short-Horizon SFT and RL} trains the base models on the original training set; \textbf{5) Pre-LLM} includes GAZP+BERT~\cite{zhong2020gazp}, HIE-SQL~\cite{zheng2022hie}, and RASAT+PICARD~\cite{qi2022rasat}, which boost SQL generation accuracy by modeling grammar, relational structures and using incremental parsing.


\subsection{RQ1: Effectiveness, Generalization and Efficiency}

\paragraph{Effectiveness: Our Framework Significantly Boosts Multi-Turn Text-to-SQL Performance, and \ul{both Warm-Start SFT and RL contribute a lot.}}
\Finding{1}{Our framework achieves the best performance compared to all previous baselines in the same model size across all datasets.}{overall_perf}. We can also observe that our method is only built based on 1.7B/4B, but achieves the best performance in both in-domain and out-of-domain settings, which even outperforms baselines with large-size models. 
We analyze the contributions from training stages (Warm-Start SFT and RL). For Warm-Start SFT, we analyze the performance and coverage of different rounds.  
\FindingSimple{2}{Self-taught Warm-Start SFT increases the coverage of high-quality long-horizon trajectories and improves downstream performance.}{overall_perf}. As the number of self-taught rounds increases, performance improves, and more training samples obtain usable trajectories (Table~\ref{tab:stc-stats-simple}).
For End-to-End RL, we observe: 
\FindingSimple{3}{RL improves both EX and EM in in-domain and out-of-domain settings.}{overall_perf,tab:inout_avgs}.
\Finding{4}{Conventional SFT attains comparable EM but exhibits weaker logical consistency; \ul{our long-horizon agent substantially improves logical correctness} (EX), which is more critical for SQL generation while maintaining or improving EM.}{overall_perf}.
\Finding{5}{Small LLMs struggle to follow long-horizon function-calling instructions}{overall_perf}. 

\begin{figure}[t] 
\vspace{-0in}
\centering \centering \includegraphics[width=\linewidth]{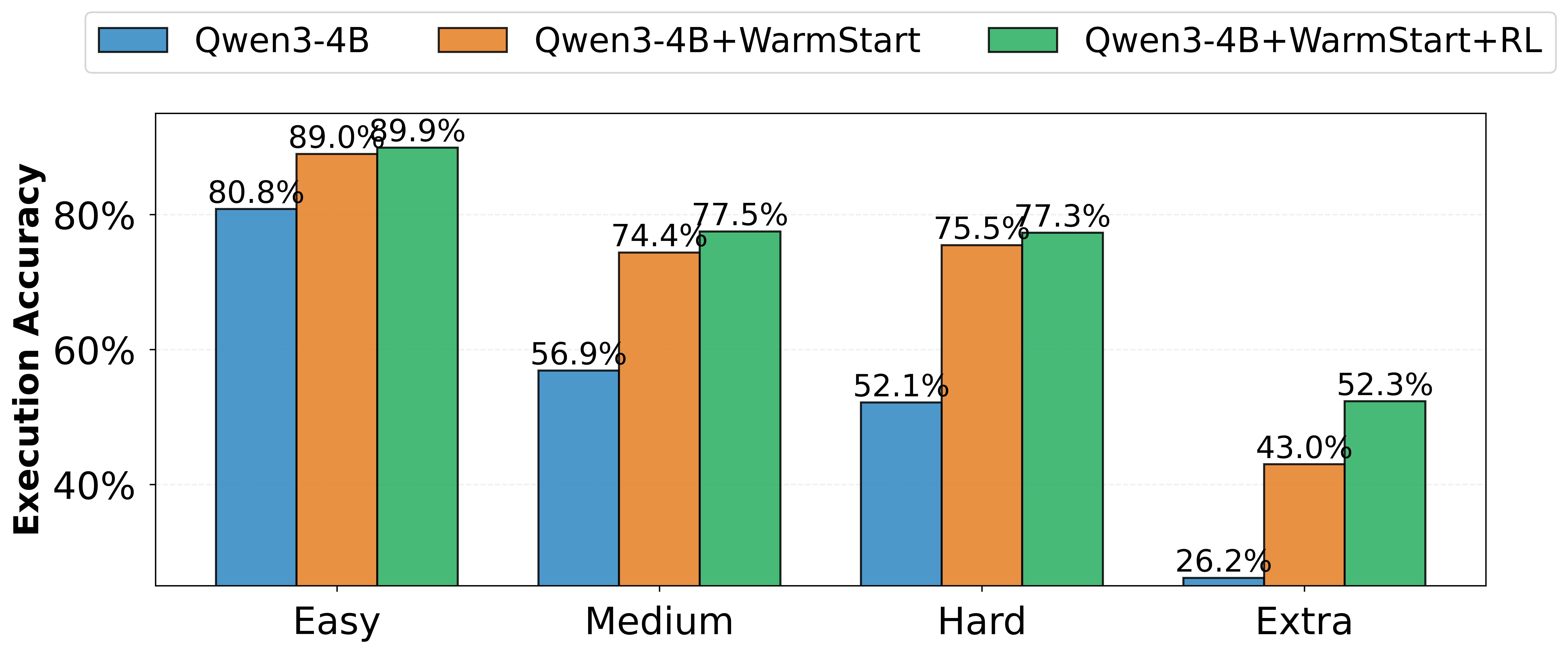} \caption{\textbf{Effectiveness: Accuracy by difficulty}. Warm-Start helps across buckets; RL further boosts performance, especially on harder queries.} \label{fig:acc-diff1} \vspace{-0.2in} 
\end{figure} 

\paragraph{Effectiveness: The long-horizon loop drives substantial improvements, while process rewards provide modest but \ul{consistent} gains, and both \ul{Execute and Memory-Verify} are essential during long-horizon reasoning.}

\FindingSimple{6}{Process Reward helps the model learn from hard examples, further boosting performance compared with sparse outcome-only rewards.}{overall_perf,fig:test_score_reward}. We begin the process reward training from medium-difficulty data. Tracking test-set scores shows larger gains on hard examples relative to outcome-only training.
For the ablations on two actions, we observe:
\FindingSimple{7}{Both \textsc{Execute} and \textsc{Memory-Verify} are essential during long-horizon reasoning.}{tab:ablations_two_actions}
Using our RL-trained model and the Qwen-14B base model on CoSQL, removing either action consistently degrades performance. 

\paragraph{Effectiveness: Our framework boosts performance across different turns and difficulty levels, with particularly strong gains on \ul{hard or deep-turn questions.}}
\FindingSimple{8}{Long-horizon reasoning yields larger gains on multi-turn dialogues and complex questions}{fig:acc-diff1,fig:diff-pair,fig:turn-pair}. 
On CoSQL, we examine accuracy across dialogue turns (Turn 1 uses no history; Turn 2 includes one prior turn, etc.) and difficulty buckets. We observe: \textbf{(i)} the base model degrades sharply as turns increase, indicating difficulty with multi-turn Text-to-SQL; \textbf{(ii)} our method improves accuracy across all turn levels, with the largest gains for Turn $\geq$ 4, highlighting the value of long-horizon modules, especially memory-based verification; \textbf{(iii)} similar patterns hold for difficulty: the base model struggles on \emph{Hard} and \emph{Extra Hard}, while our approach improves in these buckets.
\FindingSimple{9}{More difficult or more multi-turn questions require longer responses and more interactions.}{fig:diff-pair} We analyze the number of tokens and tool interactions across turn levels and difficulty buckets. The model spends more tokens as turns increase and uses more interactions on \emph{Hard}/\emph{Extra Hard} queries.

\paragraph{Generalization and Robustness: Our framework demonstrates robust generalization capabilities across \ul{out-of-domain scenarios}, \ul{different base models}, \ul{single-turn settings}, and \ul{multi-turn settings with varying history sources (ground-truth or model-generated SQL)}.}
\FindingSimple{10}{Long-horizon reasoning improves generalization across multiple dimensions.}{tab:inout_avgs,tab:generalization_across_base_models,tab:on_single_turn_datasets} \textbf{a)} As shown in Table~\ref{tab:inout_avgs}, while traditional SFT achieves strong EM in-domain, long-horizon RL substantially improves both EX and out-of-domain performance. \textbf{b)} Beyond Qwen, we applied our methods to LLaMA3.2-3B-Instruct~\cite{grattafiori2024llama}; as shown in Table~\ref{tab:generalization_across_base_models}, our method robustly improves performance across different base models. \textbf{c)} We further evaluate our models on single-turn datasets. Note that these models are trained on multi-turn datasets, making single-turn evaluation an out-of-domain test. As shown in Table~\ref{tab:on_single_turn_datasets}, this experiment indicates the robustness of our method. \textbf{d)} In our default multi-turn setting, following prior work, the dialogue history is populated with ground-truth SQL in order to isolate turn-level performance. However, relying on predicted history is arguably the only faithful measure of agentic robustness in the wild. We therefore construct a more challenging evaluation on CoSQL, where both the previous best baseline and our method must condition on model-generated prior SQL. As shown in Table~\ref{tab:robustness_predicted_prior}, our method not only continues to outperform the baseline but also \emph{widens} the performance gap (+5.3 vs.\ +4.7) when forced to rely on its own predictions, explicitly demonstrating the superior robustness of our verification-and-refinement loop.

\begin{table}[t]
\centering
\vspace{-0in}
\small
\resizebox{\linewidth}{!}{%
\begin{tabular}{lcc}
\hline
\textbf{Method} & \textbf{EX (\%)} & \textbf{EM (\%)} \\
\hline
\rowcolor{gray!12}
\textbf{Qwen3-14B (Long-horizon, no training)} & \textbf{74.4} & \textbf{55.1} \\
\rowcolor{gray!12}
\quad w/o Execution Tool & 71.4 & 54.6 \\
\rowcolor{gray!12}
\quad w/o Memory Verification Tool & 73.2 & 53.6 \\
\rowcolor{gray!12}
\quad Direct (no long-horizon reasoning) & 66.5 & 54.3 \\
\hline
\rowcolor{Qwen4B}
\textbf{Qwen3-4B + Warm-Start + RL (Ours)} & \textbf{79.9} & \textbf{65.2} \\
\rowcolor{Qwen4B}
\quad w/o Execution Tool & 74.6 & 64.6 \\
\rowcolor{Qwen4B}
\quad w/o Memory Verification Tool & 77.8 & 64.1 \\
\hline
\end{tabular}
}
\caption{Effectiveness: Ablation on two long-horizon actions (\textsc{Execute} and \textsc{Memory-Verify}).}
\label{tab:ablations_two_actions}
\vspace{-0.1in}
\end{table}

\begin{table}[t]
\centering
\vspace{-0in}
\small
\setlength{\tabcolsep}{6pt}
\renewcommand{\arraystretch}{1.15}
\begin{threeparttable}
\resizebox{\linewidth}{!}{%
\begin{tabular}{lcccc}
\toprule
& \multicolumn{2}{c}{\textbf{In-Domain (\%)}} & \multicolumn{2}{c}{\textbf{Out-of-Domain (\%)}} \\
\cmidrule(lr){2-3}\cmidrule(lr){4-5}
\textbf{Model} & \textbf{EX$\uparrow$} & \textbf{EM$\uparrow$} & \textbf{EX$\uparrow$} & \textbf{EM$\uparrow$} \\
\midrule
\rowcolor{Qwen17} Qwen3-1.7B + SFT (Short-Horizon) & 71.2 & 64.2 & 67.9 & 60.1 \\
\rowcolor{Qwen17} \textbf{Qwen3-1.7B + Warm-Start + RL (Ours) } 
& \textbf{76.8} & \textbf{64.8} & \textbf{72.5} & \textbf{63.9} \\
\midrule
\rowcolor{Qwen4B} Qwen3-4B + SFT (Short-Horizon)& 75.7 & \textbf{68.2} & 72.7 & 65.0 \\
\rowcolor{Qwen4B} \textbf{Qwen3-4B + Warm-Start + RL (Ours) } 
& \textbf{79.5} & 67.0 & \textbf{75.7} & \textbf{66.0} \\
\bottomrule
\end{tabular}
}
\vspace{-0in}
\caption{Generalization: Averaged In-domain and Out-of-domain EX/EM for the selected methods.}
\label{tab:inout_avgs}
\vspace{-0.2in}
\end{threeparttable}
\end{table}

\begin{table}[t]
\vspace{-0in}
\centering
\resizebox{1.0\linewidth}{!}{%
\begin{tabular}{lcc}
\hline
\textbf{Method} & \textbf{CoSQL (EX)} & \textbf{SParC (EX)} \\
\hline
\multicolumn{3}{l}{\textcolor{blue}{\textit{Qwen3-4B}}} \\
\rowcolor{Qwen4B} Base & 64.0 & 62.9 \\
\rowcolor{Qwen4B} Short-Horizon RL (Best Baseline) & 75.2 & 75.8 \\
\rowcolor{Qwen4B} \textbf{Ours Long-Horizon} & \textbf{79.9} & \textbf{79.0} \\
\hline
\multicolumn{3}{l}{\textcolor{purple}{\textit{LLaMA3.2-3B-Instruct}}} \\
\rowcolor{gray!12} Base & 22.9 & 24.4 \\
\rowcolor{gray!12} Short-Horizon RL (Best Baseline) & 70.4 & 70.9 \\
\rowcolor{gray!12} \textbf{Ours Long-Horizon} & \textbf{74.8} & \textbf{75.2} \\
\hline
\end{tabular}%
}
\caption{Generalization across different base models.}
\label{tab:generalization_across_base_models}
\vspace{-0.1in}
\end{table}

\begin{table}[h]
\centering
\resizebox{1.0\linewidth}{!}{%
\begin{tabular}{lcc}
\toprule
\textbf{Method} & \textbf{EX (Gold as Prior)} & \textbf{EX (Predicted as Prior)} \\
\midrule
Direct RL (Prev Best Baseline) & 75.2 & 71.2 \\
\rowcolor{Qwen4B} \textbf{Ours} & \textbf{79.9 (+4.7)} & \textbf{76.5 (+5.3)} \\
\bottomrule
\end{tabular}
}
\caption{Robustness evaluation on CoSQL under different sources of dialogue history. \textit{Gold as Prior} uses ground-truth SQL for previous turns (the standard multi-turn setting), while \textit{Predicted as Prior} uses the model's own generated SQL, reflecting realistic deployment conditions.}
\label{tab:robustness_predicted_prior}
\vspace{-0.2in}
\end{table}

\begin{figure}[t] \centering \centering \resizebox{\linewidth}{!}{\includegraphics[width=\linewidth]{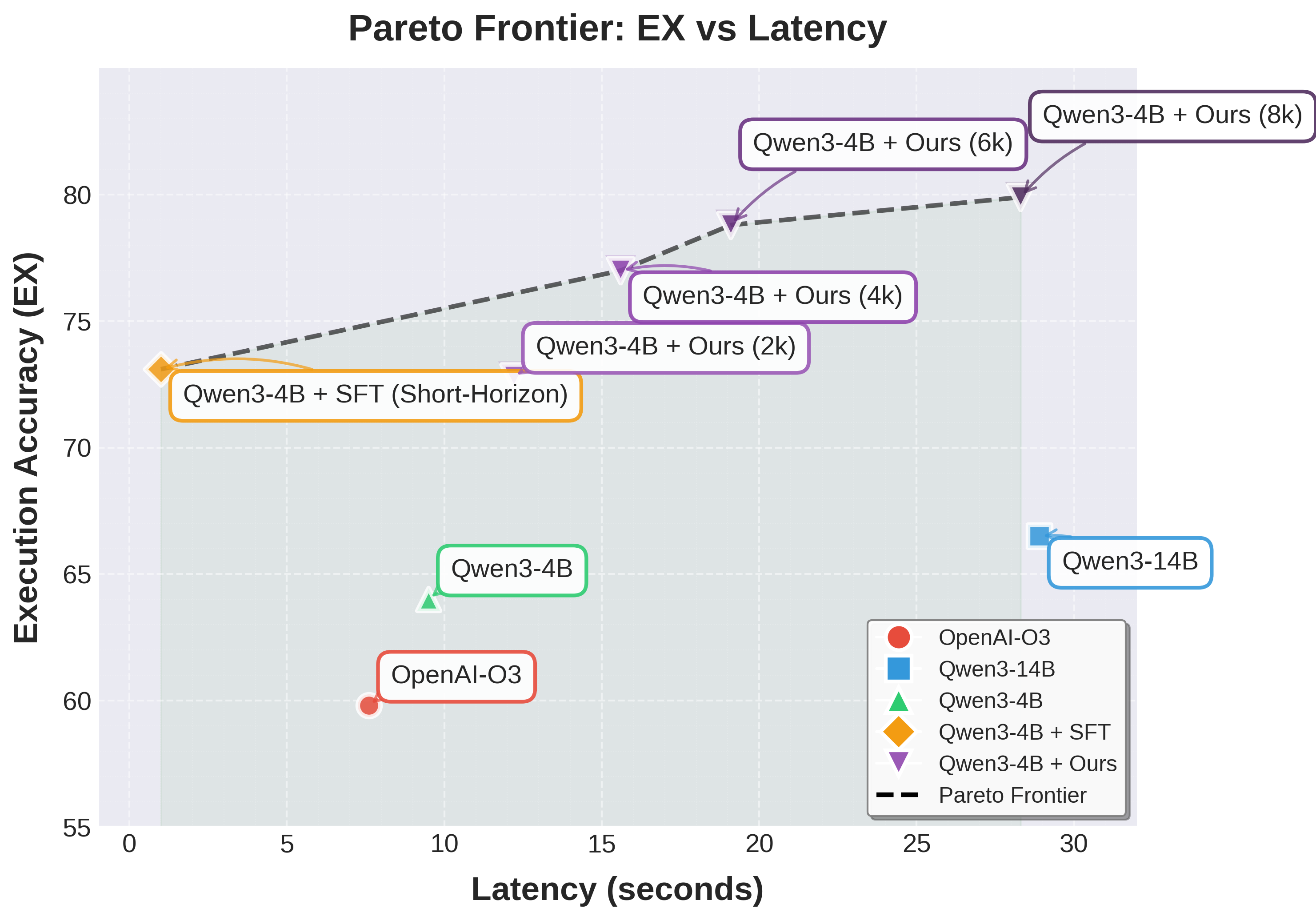} }
\vspace{-0.1in}
\caption{Efficiency-Performance Tradeoff.} \label{fig:efficiency_performance_1} \vspace{-0.1in} \end{figure}

\paragraph{Pareto Frontier: Efficiency-Performance Tradeoff.}
As shown in Fig.~\ref{fig:efficiency_performance_1}, we report execution accuracy (EX) and latency of our methods under different maximum output token limits. \ul{The results show that our approach lies on the Pareto frontier between efficiency and performance.} According to a recent survey on AI agents in practice~\cite{pan2025measuring}, around 70\% of applications tolerate latencies on the order of minutes. In comparison, our method has a maximum latency of about 28 seconds. Therefore, for high-risk enterprise data access scenarios (e.g., finance, healthcare, security logs, or batch offline processing), where incorrect SQL may expose sensitive information or cause severe business impact, our method is particularly attractive due to its substantial EX improvement. We provide detailed results in Table~\ref{tab:latency_tokens} and Fig.~\ref{fig:pareto_frontier_ex_vs_tokens} in the appendix to illustrate the trade-offs among EX, latency, and token usage. For everyday BI and interactive dashboard analytics, practitioners can select strategies that best balance performance, latency, and cost according to their requirements.


\subsection{RQ2: Quantifying Long-Horizon Abilities}

\begin{figure}[t]
\vspace{-0in}
  \centering
  \resizebox{\linewidth}{!}{%
  \includegraphics[width=\linewidth]{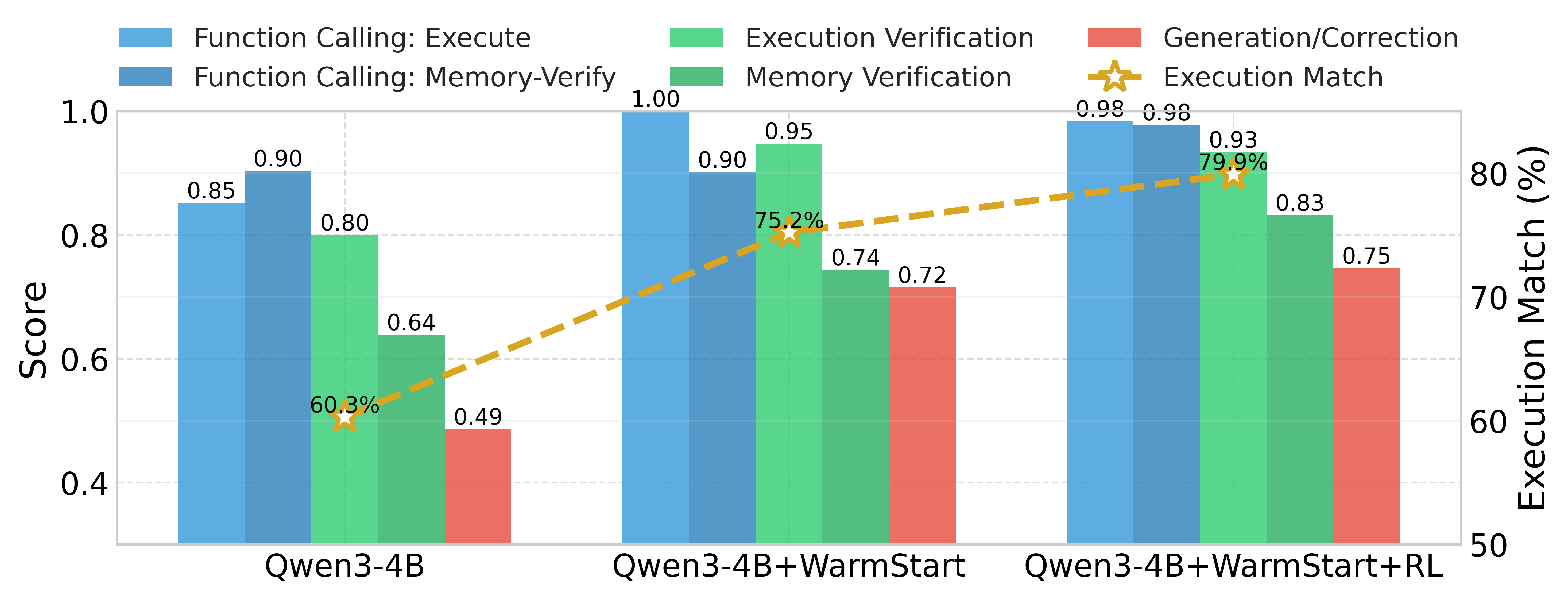}
  }
  \caption{The evolution of different Long-Horizon Abilities and related Execution Match performance from base model to RL model for Qwen3-4B.}
  \label{fig:Capabilities_4b}
  \vspace{-0.1in}
\end{figure}

We evaluate our defined five capabilities in the previous MDP: (1) function calling: \textsc{Execute} (follows tool invocation instructions), (2) function calling: \textsc{Memory-Verify}, (3) execution verification, (4) memory-based verification, and (5) generation/self-correction. For (1)-(2), a trial is successful if the prescribed tools are invoked; otherwise, it scores zero. For (3)--(5), we use the process rewards defined earlier. We also track execution accuracy to relate these abilities to overall performance, for 1.7B and 4B models across three stages: Base, Warm-Start, and Warm-Start+RL.
As shown in Fig.~\ref{fig:Capabilities_4b}, ~\ref{fig:perf-17b}, we observe: \ul{\textbf{(i)} all five abilities improve with Warm-Start and further with RL; \textbf{(ii)} RL especially boosts memory-related abilities (both calling and verification); \textbf{(iii)} The 1.7B base model is much weaker than the 4B model primarily due to weaker long-horizon abilities, but both benefit from our training.}

\paragraph{Correlation with Overall Accuracy}
\FindingSimple{11}{Stronger function calling, verification, and self-correction correlate with better SQL performance.}{fig:Capabilities_4b}

\subsection{RQ3: SQL Generation Quality}
\paragraph{Which SQL errors are mitigated?}

\begin{figure}[h]
  \centering
  \includegraphics[width=0.85\linewidth]{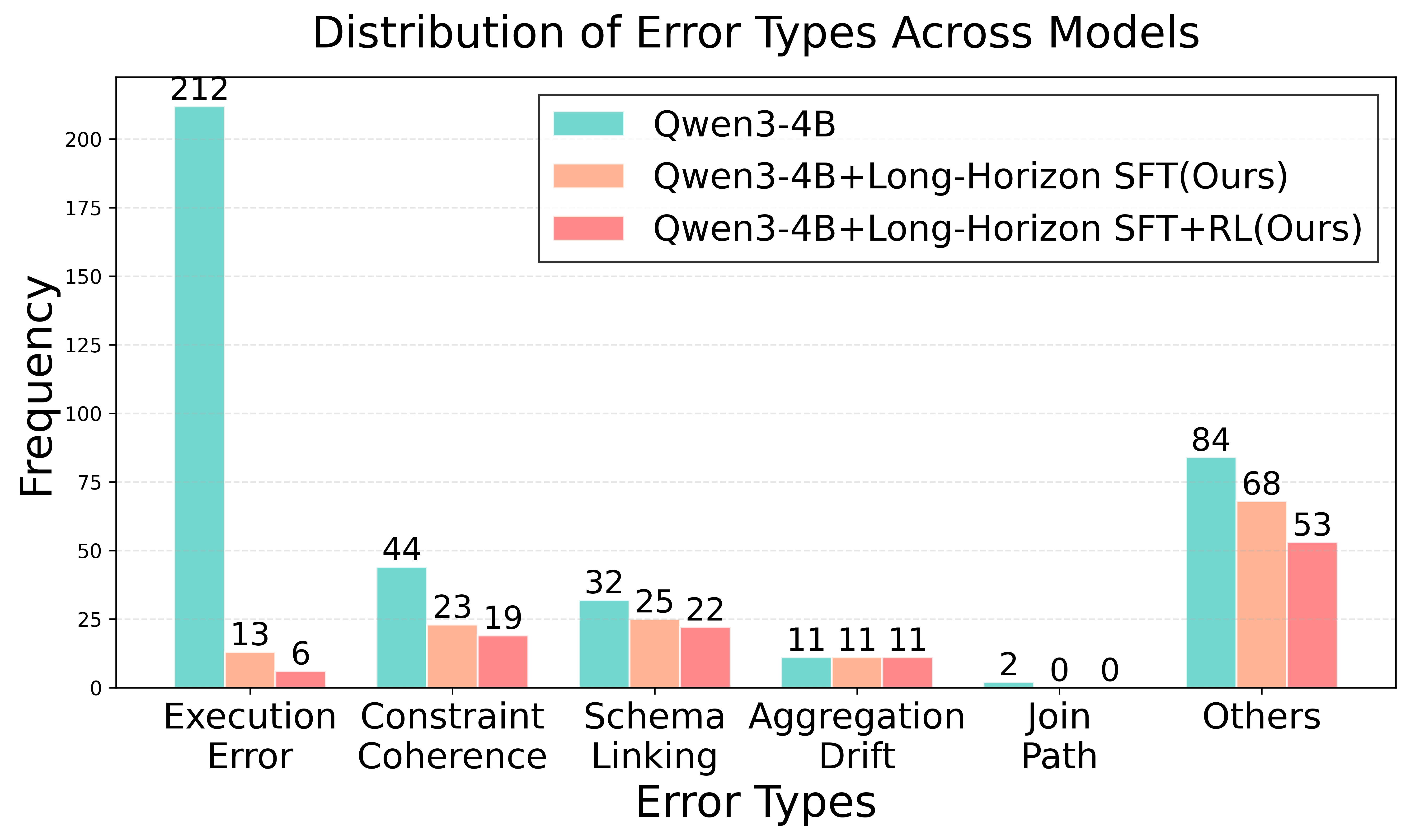} 
   \vspace{-0in}
  \caption{Distribution of error types across models.}
  \label{fig:sql_error}
\end{figure}

We adopt our designed error taxonomy, \emph{Execution Error} plus four coherence-related errors: \emph{Constraint Coherence}, \emph{Schema Linking}, \emph{Aggregation Drift}, and \emph{Join Path}, and use an LLM-as-judge approach using GPT-5 as the judge (given ground truth, prediction, and dialogue history) to assess error incidence before/after training. From Fig.~\ref{fig:sql_error}, we find: \ul{\textbf{(i)} Execution errors drop sharply}, consistent with adding execution and verification actions; note that six of the remaining execution failures stem from an 8,000-token cap (truncation before completion); \ul{\textbf{(ii)} context-coherence errors (Constraint Coherence, Schema Linking, Join Path) decrease substantially}, indicating stronger context adherence and verification; \ul{\textbf{(iii)} Aggregation Drift changes little}, since aggregation drift-related SQL are mostly extra hard, suggesting a hard open problem on extra-hard queries and a direction for future work.


\vspace{-0.1in}

\paragraph{Case Studies.}
\FindingSimple{12}{Through long-horizon training, the agent learns to resolve execution failures (including null-return cases, i.e., \textbf{aha-moments} in Text-to-SQL) and coherence errors.}{fig:SFT,pg:Case2} Key reasoning is highlighted.

%% file: latex/5-Conclusion.tex
\section{Discussion}
\label{sec:Discussion}

We highlight two practical considerations beyond our core experimental scope of this work: \textbf{scaling to industrial-scale schemas} and \textbf{the flexibility of our memory design}. Because our method is an agentic tool-use framework, it extends naturally to industrial-scale schemas where loading the full schema into context is infeasible. Scalability can be addressed by introducing a \texttt{Schema\_Retrieval} tool invoked prior to the \texttt{PROPOSE} step: instead of loading the entire schema, the agent queries a vector database of table and column descriptions to retrieve only the top-$K$ relevant tables. Beyond schema scalability, our memory module is also designed to be backend-agnostic and compatible with both structured and unstructured representations. The agent interacts with memory exclusively through function calls (e.g., \texttt{memory\_retrieve}), decoupling the backend from the LLM. While the current implementation stores parsed constraints as text, the backend can be transparently upgraded to a vector database for long dialogues or unstructured guidelines, with \texttt{memory\_retrieve} performing similarity search over historical constraints.

\section{Conclusion}
\label{sec:Conclusion}



In this work, we propose \textsc{MTSQL-R1}, the first multi-turn Text-to-SQL agent trained with explicit long-horizon reasoning. Experiments on CoSQL and SParC show that \textsc{MTSQL-R1} outperforms all baselines, highlighting the value of long-horizon reasoning for conversational semantic parsing and its potential for future research.

%% file: latex/Appendix.tex
\clearpage
\appendix
\startcontents[app]

\printcontents[app]{}{1}{\section*{Appendix Contents}}

\begin{figure}[!htbp]
  \centering
\includegraphics[width=1\linewidth]{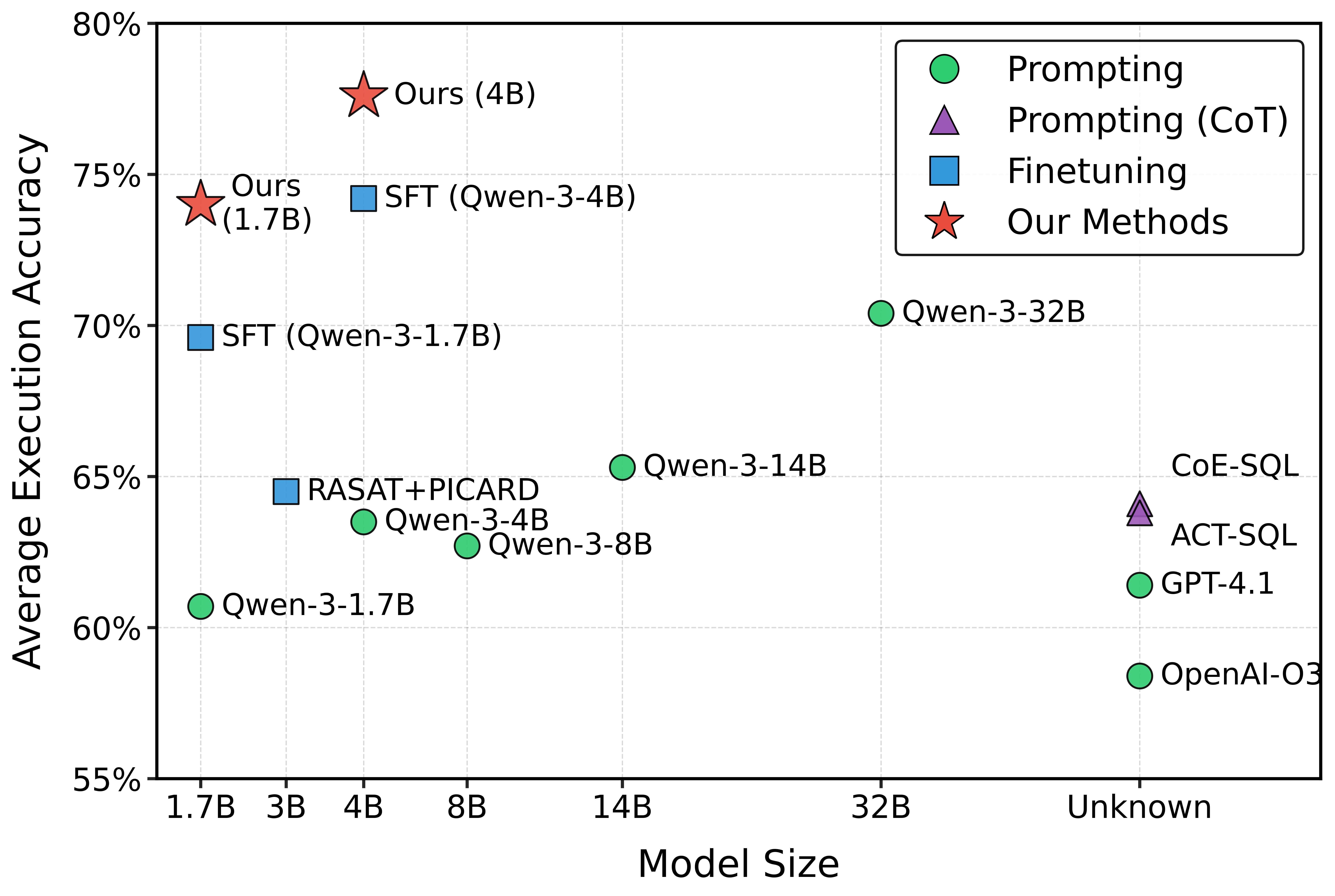} 
  \caption{Comparison between existing methods and our MTSQL-R1 on the average of CoSQL and SParC benchmark. Our method outperforms both strong prompting-based
and finetuned baselines, achieving superior performance
across various model sizes.}
  \label{fig:overview_position}
\end{figure}

\section{The detailed Algorithm of Self-Taught Warm-Start SFT}

The detailed Algorithm of Self-Taught Warm-Start SFT is shown in \Cref{ws_sft}.

\begin{algorithm}[t]
\tiny              
\caption{Self-Taught Warm-Start SFT}
\label{ws_sft}
\KwIn{Policy $\pi_{\theta_0}$, data $D_0=\{(I,y^\ast)\}$, rounds $N$.}
\KwOut{Policy $\pi_{\theta^\star}$.}
$\Traj\leftarrow\emptyset$\;
\For{$i=0$ \KwTo $N-1$}{
  \Stage{S1: Collect 20 rollouts per item with temp 0.7}
  $\Traj_i^{\text{raw}}\!\leftarrow\!\bigcup_{I\in D_i}\{(I,a_{1:n},\hat y_{1:n})\sim\pi_{\theta_i}(\cdot|I)\}$\;
  $\Traj_i^{\text{valid}}\!\leftarrow\!\{\tau\in\Traj_i^{\text{raw}}~|~\EM(\hat y_n,y^\ast)\land\EX(\hat y_n,y^\ast)\}$\;

  \Stage{S2: Difficulty-aware rejection sampling}
  
  \ForEach{$I$ with $\tau\in\Traj_i^{\text{valid}}$}{
    \eIf{$I$ is (Easy-SQL \textbf{or} 20/20 correct)}{
      keep less-interaction trajectories; sample up to $2$\;
    }{
      keep long-interaction trajectories; sample $3$ after clustering by embedding \;
    }
    add sampled trajectories to $\Traj$\;
  }

  \Stage{S3: Supervised fine-tuning}
  $\pi_{\theta_{i+1}}\leftarrow \operatorname{SFT}(\pi_{\theta_i},\Traj)$\;

  \Stage{S4: Update data}
  $D_{i+1}\leftarrow D_i\setminus\{I~|~\exists\tau\in\Traj_i^{\text{valid}}\text{ for }I\}$\;
}
\Return{$\pi_{\theta_{i+1}}$}\;
\end{algorithm}

\section{Dataset Statistics: The SQL Hardness Criteria and Statistics of Two Multiturn Text-to-SQL Datasets} We follow the previous method in~\cite{yu-etal-2018-spider} to divide SQL queries into 4 levels: easy, medium, hard, extra hard.  We grade query difficulty by counting SQL elements, especially selections and conditions. Queries that use more SQL constructs (e.g., \texttt{GROUP BY}, \texttt{ORDER BY}, set operations such as \texttt{INTERSECT}, nested subqueries, multiple column selections, and aggregators) are treated as harder. Concretely, a query is labeled \textbf{hard} if it has more than two selected columns, more than two \texttt{WHERE} predicates, and a \texttt{GROUP BY} on two columns, or if it includes \texttt{EXCEPT} or nesting. Queries that add further complexity beyond these thresholds are labeled \textbf{extra hard}.

For details, please see Table \ref{fig:data_statistics} and Table. \ref{fig:data_diff}.

\begin{table}[!ht]
\centering
\begin{tabular}{lcc}
\toprule
 & \textbf{CoSQL} & \textbf{SParC} \\
\midrule
\# Q sequences        & 3,007          & 4,298 \\
\# user questions     & 15,598   & 12,726 \\
\# databases          & 200            & 200 \\
\# tables             & 1,020          & 1,020 \\
Avg. Question Length           & 11.2           & 8.1 \\
Vocab                 & 9,585          & 3,794 \\
Avg. Turns      & 5.2            & 3.0 \\
Unanswerable Q        & \ding{51}      & \ding{55} \\
User intent           & \ding{51}      & \ding{55} \\
System response       & \ding{51}      & \ding{55} \\
\bottomrule
\end{tabular}
\caption{Dataset comparison between CoSQL and SParC.}
  \label{fig:data_statistics}
\end{table}

\begin{table}[!ht]
\centering
\resizebox{0.48\textwidth}{!}{%
\begin{tabular}{lcccc}
\toprule
\textbf{Dataset} & \textbf{Easy} & \textbf{Medium} & \textbf{Hard} & \textbf{Extra Hard} \\
\midrule
SParC  & 40.1\% & 36.7\% & 12.1\% & 11.1\% \\
CoSQL  & 41.4\% & 31.8\% & \textbf{16.2\%} & 10.5\% \\
\bottomrule
\end{tabular}
}
\caption{Difficulty distribution by dataset.}
\label{fig:data_diff}
\end{table}

\section{Implementation Details}
\label{sec:Implementation_details}

We implement our method using the latest open-source reasoning model Qwen3-1.7B and Qwen3-4B~\cite{yang2025qwen3technicalreport} as the backbone model. Our models are trained on a single node of 8 NVIDIA A100 GPUs.  
For Self-Taught Warm-Start SFT, we use LlamaFactory~\cite{zheng-etal-2024-llamafactory}, which adopts DeepSpeed~\cite{aminabadi2022deepspeedinferenceenablingefficient} for distributed training with ZeRO-3 offload,
along with gradient checkpointing. we use a learning rate of 5e-6, a cosine learning rate scheduler, a per-device training batch size of 2, and full parameter fine-tuning.
For End-to-End GRPO Training, we utilize the GRPO implementation from the Verl package~\cite{sheng2024hybridflow} with FSDP parameter offloading enabled and SGLang~\cite{zheng2024sglangefficientexecutionstructured} as the inference engine. The training batch size is set to 256, the maximum prompt length is 4000, and the maximum response length is 8000. The learning rate is 1e-6, the maximum interaction between agent and tools is set to 4, and the number of rollouts is 5.

\subsection{Long-Horizon Reasoning as Tools Settings}

\paragraph{Tool Description Configuration}

For interacting with the database, we have the ``exec\_sql'' tool:
\begin{tooldesc}
- class_name: "verl.tools.text2sql_tool.Text2sqlTool"
config: {}
tool_schema:
  type: "function"
  function:
    name: "exec_sql"
    description: "A tool for executing sql and return the query results"
    parameters:
      type: "object"
      properties:
        code:
          type: "string"
          description: "The current generated SQL that will be executed"
      required: ["code"]
\end{tooldesc}

The return message of the ``exec\_sql'' tool is:
\begin{tooldesc}
Recap:  
- Current question: {current_q}  
- Generated SQL: {code}  
- SQL execution results (truncated to 200 characters): {return_msg}  

Now please:  
1. Verify whether the SQL execution results are valid:  
   - Check if the SQL runs without errors.  
   - Check if the returned columns exist in the schema and are relevant to the question.  
   - Check if the results contain unexpected NULL values, empty sets, or error messages.  

2. After verifying, output:  
   - <exec_verify>pass</exec_verify> if the results are valid and consistent with the schema.  
   - <exec_verify>no_pass</exec_verify> if the results show errors, irrelevant columns, or invalid values.  

3. If <exec_verify>no_pass</exec_verify>, think step by step, refine the SQL and provide a corrected SQL and then execute it via re-calling ``exec_sql`` tool again via <tool_call>. Repeat until you get valid results.
4. If <exec_verify>pass</exec_verify>, You have to call `memory_retrieve` tool via <tool_call>  at least once to ensure the current generated SQL is coherent with the historical memory.
\end{tooldesc}

For interacting with memory, we have the ``"memory\_retrieve" tool:

\begin{tooldesc}
- class_name: "verl.tools.memory_retriever.MemoryRetriever"
config: {}
tool_schema:
type: "function"
function:
name: "memory_retrieve"
description: "A tool for retrieving the historical questions and ground-truth SQL in this dialogue"
parameters:
  type: "object"
  properties:
    code:
      type: "string"
      description: "The current generated SQL that needs to be verified coherence with the given historical memory"
  required: [ "code" ]
\end{tooldesc}

The return message of the ``"memory\_retrieve" tool is:

\begin{tooldesc}
You are a coherence verifier for Multi-turn Text2SQL.

Current Question: {current_q}  
Proposed SQL: {code}  
The execution results of the proposed SQL: {execution_results}

Memory (historical information in order):  
{memory_str}  

Your tasks:  
1. Verify whether the Proposed SQL is coherent with the Current Question and the Memory, based on the relation between the Current Question and Historical Questions.  
   - If the Current Question introduces changes (new columns, conditions, ordering, etc.), SQL should update accordingly.  
   - If not, SQL must remain consistent with the Historical Questions.  

Step-by-step reasoning checklist:  
   1. First parse the Proposed SQL into its components (SELECT, FROM, WHERE, GROUP BY, HAVING, ORDER BY, JOINs).
   2. Check tables are consistent with context.  
   3. Check selected columns match current and historical intent.  
   4. Check conditions (WHERE/GROUP/HAVING) reflect the relation between current and past questions.  
   5. Check ordering (ORDER BY) is preserved unless explicitly changed.  
   6. Verify that joins and table relationships follow the established context.
   7. Make sure the SQL and the execution results of the proposed SQL answer the current question while remaining logically coherent with the conversation history and execution results.

2. After verifying, output one of the following:  
   - `<memory_verify>pass</memory_verify>` if coherent.  
   - `<memory_verify>no_pass</memory_verify>` if not coherent.  

3. If `no_pass`: explain issues, think step by step to refine SQL, and then please call `exec_sql` tool again via <tool_call> to check the corrected SQL and get the execution results. Repeat until you get `pass`. 
4. If `pass`: return the final SQL inside `<answer_sql>...</answer_sql>`.  

Note finally you should return the final SQL inside `<answer_sql>...</answer_sql>
\end{tooldesc}

\paragraph{Tool-Related Hyperparameters}

\begin{tooldesc}
actor_rollout_ref:
hybrid_engine: True
rollout:
name: sglang
multi_turn:
  enable: True
  max_turns: 4   # Important Max-turns
\end{tooldesc}

\subsection{Hyperparameter Settings}

\begin{hps}
    config-name='text2sql_multiturn_grpo' \
custom_reward_function.path=verl/utils/reward_score/text2sql_process.py  \
    algorithm.adv_estimator=grpo \
    data.train_files=train_rl{DATA_LABEL}.parquet \
    data.val_files=test.parquet \
    data.train_batch_size=256 \
    data.max_prompt_length=4000 \
    data.max_response_length=8000 \
    data.filter_overlong_prompts=True \
    data.truncation='error' \
    data.return_raw_chat=True \
    actor_rollout_ref.model.path=MODEL_PATH \
    actor_rollout_ref.actor.optim.lr=1e-6 \
    actor_rollout_ref.model.use_remove_padding=True \
    actor_rollout_ref.actor.ppo_mini_batch_size=256 \
    actor_rollout_ref.actor.ppo_micro_batch_size_per_gpu=32 \
    actor_rollout_ref.actor.use_kl_loss=False \  
    actor_rollout_ref.actor.kl_loss_coef=0.001 \
    actor_rollout_ref.actor.kl_loss_type=low_var_kl \
    actor_rollout_ref.actor.entropy_coeff=0 \
    actor_rollout_ref.model.use_fused_kernels=True \
    actor_rollout_ref.actor.use_dynamic_bsz=True \
    actor_rollout_ref.actor.ppo_max_token_len_per_gpu=30000 \
    actor_rollout_ref.rollout.log_prob_use_dynamic_bsz=true \
    actor_rollout_ref.rollout.log_prob_max_token_len_per_gpu=34000 \
    actor_rollout_ref.ref.log_prob_use_dynamic_bsz=true  \
    actor_rollout_ref.ref.log_prob_max_token_len_per_gpu=34000 \
    actor_rollout_ref.model.enable_gradient_checkpointing=True \
    actor_rollout_ref.actor.fsdp_config.param_offload=False \
    actor_rollout_ref.actor.fsdp_config.optimizer_offload=False \
    actor_rollout_ref.rollout.log_prob_micro_batch_size_per_gpu=64 \
    actor_rollout_ref.rollout.tensor_model_parallel_size=1 \
    actor_rollout_ref.rollout.name=sglang \
    actor_rollout_ref.rollout.gpu_memory_utilization=0.8 \
    actor_rollout_ref.rollout.n=5 \
    actor_rollout_ref.ref.log_prob_micro_batch_size_per_gpu=64 \
    actor_rollout_ref.ref.fsdp_config.param_offload=True \
    algorithm.use_kl_in_reward=False \
    trainer.critic_warmup=0 \
    trainer.logger=['console','wandb'] \
    trainer.project_name='verl_grpo_text2sql' \
    trainer.experiment_name="${data}_${tag}" \
    trainer.val_before_train=True \
    trainer.n_gpus_per_node=8 \
    trainer.nnodes=1 \
    trainer.save_freq=10 \
    trainer.test_freq=10 \
    trainer.validation_data_dir="./${data}_${tag}_rollouts_sql_train/" \
actor_rollout_ref.rollout.multi_turn.tool_config_path="text2sql_tool_config.yaml" \
    trainer.total_epochs=60 \
\end{hps}

\subsection{Discussion about the training pipelines}

A frequent industry pipeline is to RL-post-train a very large model and then use it to generate SFT/distillation data, yielding a stronger initial checkpoint for training smaller models. We have not explored such pipelines due to computational constraints, but our MDP formulation is orthogonal to this strategy and can be naturally integrated into it. Notably, our self-taught SFT-then-RL pipeline trains stably even when the initial model lacks frontier-level reasoning, and our core contribution, the long-horizon multi-turn formulation, is expected to compound with pipelines that rely on RL tuning of very large models. We leave a large-scale industrial study of this direction to future work.

\section{Additional Experiments}

\subsection{Warm-Start SFT: Verified Long-Horizon Trajectories Coverage}

The increasing coverage of training examples during Self-Taught Warm-Start SFT is shown in Table~\ref{tab:stc-stats-simple}.

\begin{table}[!ht]
  \centering
  \resizebox{\columnwidth}{!}{%
    \begin{tabular}{lrr}
      \toprule
      & \textbf{CoSQL} & \textbf{SParC} \\
      \midrule
      All training examples                               &  9,337 & 11,905 \\
      Training samples that have trajectories (Round1)              &  6,311 &  9,132 \\
      Training samples that have trajectories (Round2)                                              &  7,409 & 10,103 \\
      Training samples that have trajectories (Round3)                                              &  7,555 & 10,285 \\
      Final Long-horizon Trajectories (Round 3)                  & 19,416 & 29,710 \\
      \bottomrule
    \end{tabular}%
  }
    \caption{Self-Taught Coverage Statistics (CoSQL/SParC): As self-taught rounds increase, the model strengthens and covers a larger share of training samples, yielding more high-quality, natural trajectories for Warm-Start.}
  \label{tab:stc-stats-simple}
\end{table}






\subsection{Effectiveness: Difficulty-wise and Turn-wise results: execution accuracy and token length/interactions}

\begin{figure*}[t]
  \vspace{-0.6in}
  \centering
  \begin{subfigure}{0.44\textwidth}
    \centering
    \includegraphics[width=\linewidth]{imgs/execution_accuracy_by_difficulty.png}
    \caption{\textbf{Accuracy by difficulty (Easy, Medium, Hard, Extra)}. Warm-Start helps across buckets; RL further boosts performance, especially on harder queries.}
    \label{fig:acc-diff}
  \end{subfigure}\hfill
  \begin{subfigure}{0.55\textwidth}
    \centering
    \includegraphics[width=\linewidth]{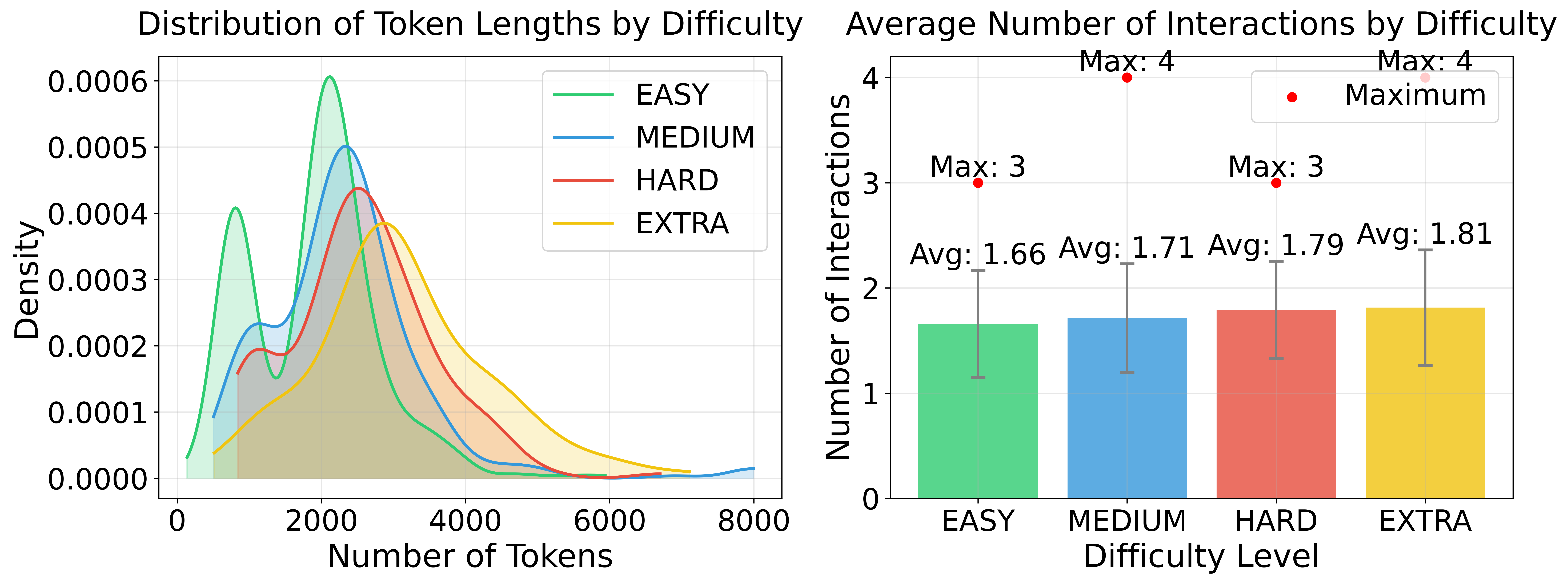}
    \caption{\textbf{Token length \& interactions by difficulty}. Harder buckets yield longer sequences and slightly more interactions.}
    \label{fig:token-diff}
  \end{subfigure}
  \caption{Difficulty-wise results: execution accuracy (\subref{fig:acc-diff}) and token length/interactions (\subref{fig:token-diff}) on CoSQL.}
  \label{fig:diff-pair}
\vspace{-0in}
\end{figure*}

\begin{figure}[H]
  \centering
  \begin{subfigure}{0.5\textwidth}
    \centering
    \includegraphics[width=\textwidth]{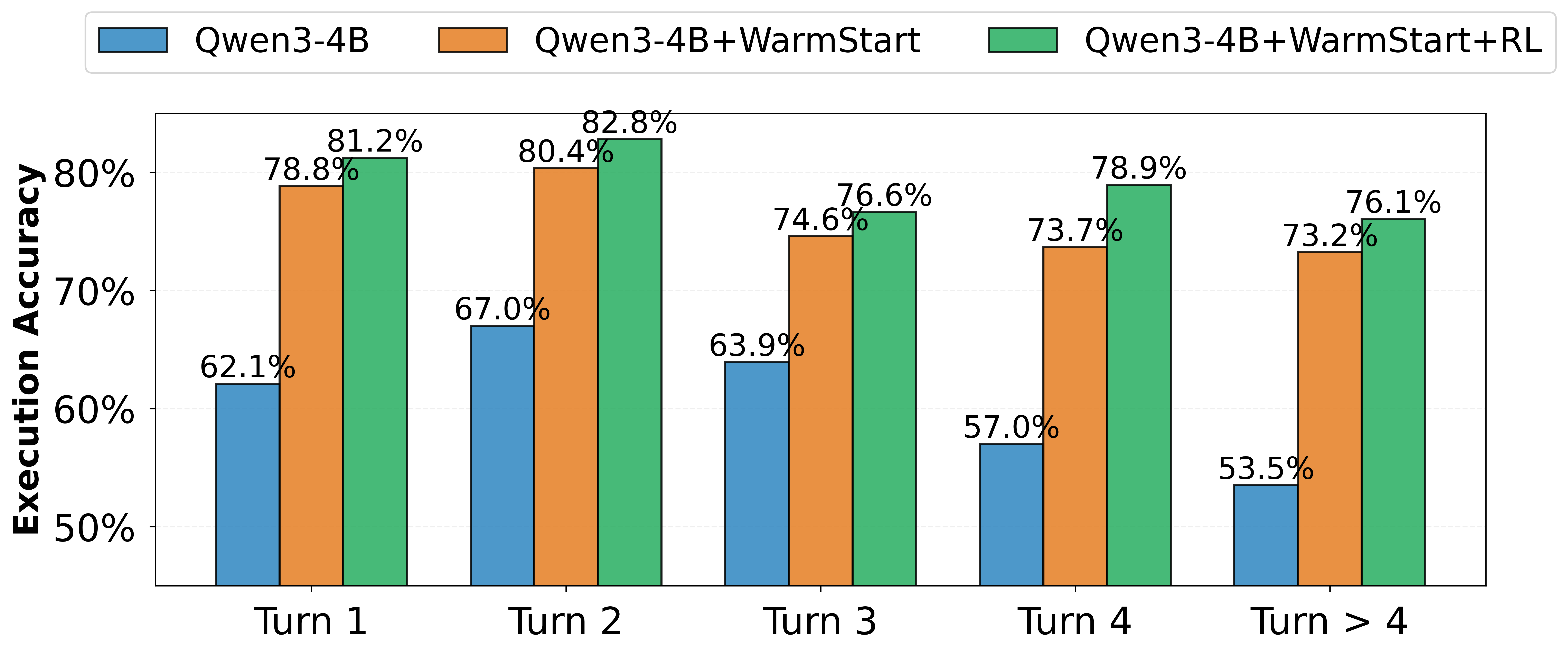}
    \caption{\textbf{Accuracy by dialogue turn (1 → >4)}. Warm-Start improves; RL yields the best results with larger gains at later turns.}
    \label{fig:acc-turn}
  \end{subfigure}\vfill
  \begin{subfigure}{0.5\textwidth}
    \centering
    \includegraphics[width=\textwidth]{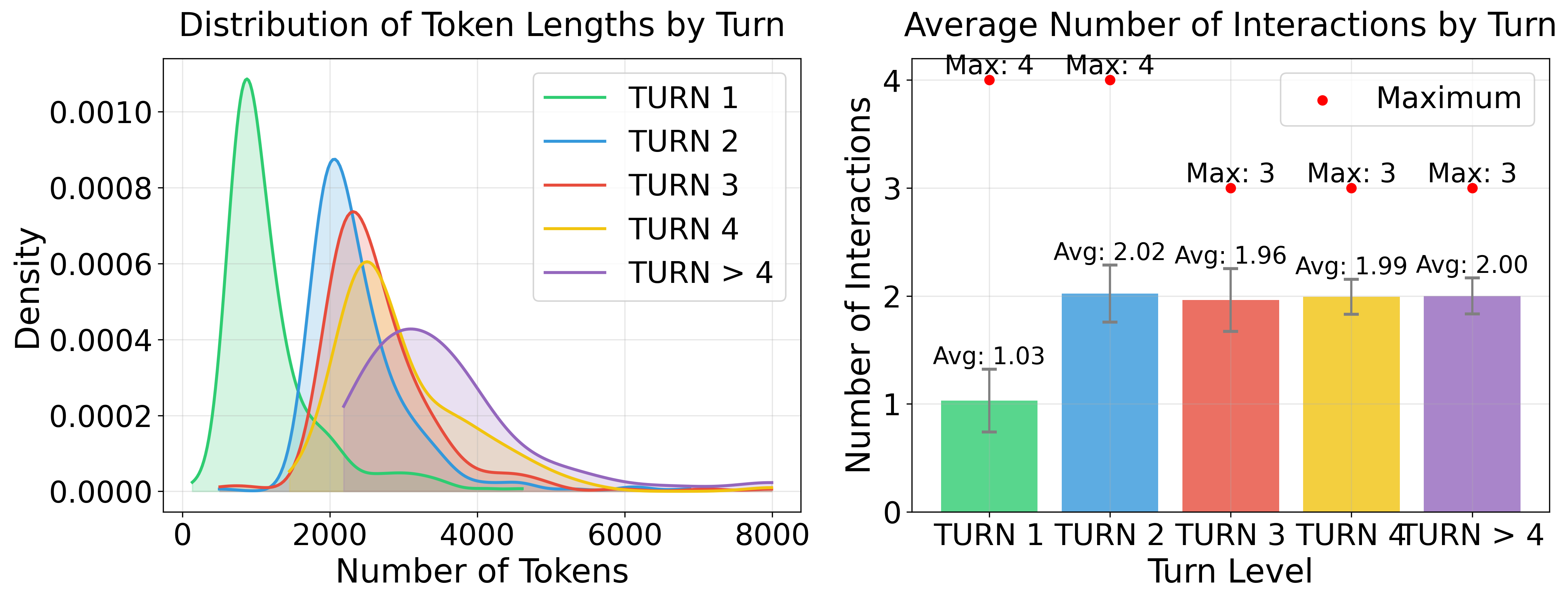}
    \caption{\textbf{Token length \& interactions by turn}. Distributions shift right and broaden as turns increase.}
    \label{fig:token-turn}
  \end{subfigure}
  \caption{CoSQL: turn-wise results: execution accuracy (\subref{fig:acc-turn}) and token length/interactions (\subref{fig:token-turn}).}
  \label{fig:turn-pair}
\end{figure}

The Difficulty-wise  accuracy and token length/interactions is shown in Fig.~\ref{fig:diff-pair}; while the Turn-wise results are shown in Fig.~\ref{fig:turn-pair}.

\subsection{Effectiveness: RL Reward Ablation Studies}

We conduct a reward ablation study to isolate the effect of each reward component using the CoSQL Qwen3-4B experimental setting. Table~\ref{tab:reward_ablation_studies} shows the results demonstrate that both reward modules clearly contribute to performance gains.

\begin{table}[t]
\centering
\resizebox{\columnwidth}{!}{%
\begin{tabular}{lcc}
\toprule
Reward in RL & EX & EM \\
\midrule
OutCome Only & $79.1 \pm 0.15$ & $64.5 \pm 0.18$ \\
Outcome + Verify Reward & $79.7 \pm 0.14$ & $65.0 \pm 0.19$ \\
Outcome + Propose/Correction Reward & $79.4 \pm 0.11$ & $65.4 \pm 0.18$ \\
\textbf{Outcome + Propose/Correction Reward + Verify} & \textbf{$79.9 \pm 0.11$} & \textbf{$65.2 \pm 0.17$} \\
\bottomrule
\end{tabular}
}
\caption{Reward Ablation Studies: The effect of each reward component on EX and EM. The results demonstrate that both reward modules clearly contribute to performance gains.}
\label{tab:reward_ablation_studies}
\end{table}

\subsection{Generalization: Full Performance Across Different Base Models}

Table~\ref{tab:generalization_across_base_models_2} presents the comprehensive performance of our methods applied to two different base models, demonstrating the robustness and generalizability of our approach across different architectures and datasets.

\begin{table*}[]
\resizebox{\linewidth}{!}{%
\begin{tabular}{lccll}
\hline
                                                        & \multicolumn{2}{l}{CoSQL}     & \multicolumn{2}{l}{SParC}                                         \\ \hline
\textbf{Method}                                         & \textbf{EX}   & \textbf{EM}   & \multicolumn{1}{c}{\textbf{EX}} & \multicolumn{1}{c}{\textbf{EM}} \\ \hline
Qwen3-4B Base                                           & 64.0          & 50.7          & 62.9                            & 49.8                            \\
Qwen3-4B + Short-Horizon RL (Best Baseline)             & 75.2          & 64.8          & 75.8                            & 66.5                            \\
\textbf{Qwen3-4B + Ours Long-Horizon}                   & \textbf{79.9} & \textbf{65.2} & \textbf{79.0}                   & \textbf{68.7}                   \\ \hline
LLaMA3.2-3B-Instruct Base                               & 22.9          & 14.6          & 24.4                            & 13.5                            \\
LLaMA3.2-3B-Instruct + Short-Horizon RL (Best Baseline) & 70.4          & 62.0          & 70.9                            & 63.2                            \\
\textbf{LLaMA3.2-3B-Instruct + Ours Long-Horizon}       & \textbf{74.8} & \textbf{63.1} & \textbf{75.2}                   & \textbf{65.0}                   \\ \hline
\end{tabular}
}
\caption{Generalization: Full Performance Across Different Base Models. Our long-horizon reasoning framework consistently outperforms both base models and short-horizon RL baselines.}
\label{tab:generalization_across_base_models_2}
\end{table*}

\subsection{Generalization: Performance of Our Multi-Turn Models on Out-of-Domain (OOD) Single-Turn Text-to-SQL Datasets}

We emphasize that both CoSQL and SParC already include single-turn evaluation (i.e., the first turn), and our method has already demonstrated superior performance in the single-turn setting.

Furthermore, to more clearly demonstrate that our trained agent also performs well on existing single-turn datasets, we applied our trained model, which was trained only on CoSQL and not on the training sets of Spider or BIRD, as a simple proof-of-concept. The results are summarized in Table~\ref{tab:on_single_turn_datasets} and indicate that our method continues to achieve excellent performance in the single-turn setting, especially given the relatively small model size, while also exhibiting strong generalization ability.

\begin{table*}[t]
\centering
\resizebox{\linewidth}{!}{%
\begin{tabular}{l l c c c c}
\toprule
\textbf{} & \textbf{Model Size} & \textbf{Use specific training set?} &
\multicolumn{1}{c}{\textbf{Spider}} &
\multicolumn{1}{c}{\textbf{BIRD}} &
\textbf{Support Multi-Turn Text-to-SQL?} \\
\cmidrule(lr){4-4}\cmidrule(lr){5-5}
 &  &  & \textbf{EX} & \textbf{EX} &  \\
\midrule
DAIL-SQL + GPT-4~\cite{gao2023text}          & Closed-Source (very large) & YES & 86.2 & 54.76 & NO \\
CHASE-SQL + Gemini 1.5~\cite{pourreza2024chase}    & Closed-Source (very large) & YES & 87.6 & 73.00 & NO \\
RESDSQL-3B + NatSQ~\cite{li2023resdsql}         & 3B                         & YES & 78.0 & --    & NO \\
T5-3B                      & 3B                         & YES & --   & 23.3  & NO \\
SFT CodeS-7B               & 7B                         & YES & --   & 57.1  & NO \\
Qwen3-4B                   & 4B                         & NO  & 70.7 & 25.8  & YES \\
Qwen3-14B                  & 14B                        & NO  & 72.9 & 32.6  & YES \\
DeepSeek                   & 236B                       & NO  & --   & 56.1  & YES \\
\midrule
Qwen3-4B + Long-Horizon Training (Ours)            & 4B                         & NO  & 84.2 & 56.9  & YES \\
\bottomrule
\end{tabular}%
}
\caption{Performance comparison on single-turn Text-to-SQL datasets. Our method achieves competitive results without training on domain-specific datasets.}
\label{tab:on_single_turn_datasets}
\end{table*}

\subsection{Efficiency-Performance Tradeoff}
\label{sec:token_len_latency}

We conducted further experiments to study how performance changes under different maximum token cost constraints during LLM inference. Using these results, we can better understand the trade-off between effectiveness and efficiency. The following Table~\ref{tab:latency_tokens} presents results for Qwen3-4B on the CoSQL dataset.

Note that for Qwen3-4B + Ours (max output = 2000 tokens), performance drops drastically, primarily because our method is designed for long-horizon reasoning. With a 2000-token cap, the agent often cannot complete its full reasoning process, so we must extract intermediate SQL before a final query is produced. This performance drop is expected, as our approach is specialized for long-horizon reasoning rather than heavily truncated outputs.

One additional point worth mentioning is that the current latency is measured using the basic version of vLLM, without explicit KV-cache optimization, quantization (using fp16, etc.), or speculative decoding for LLM agent inference. Therefore, the reported latency should be viewed as a reference value, and we expect that applying these techniques could substantially reduce latency.

\begin{table*}[t]
\centering
\begin{tabularx}{0.9\linewidth}{@{}Xrrr@{}}
\toprule
Method & EX & Latency (s) & Avg. Token generated by LLM \\
\midrule
OpenAI-O3 & 59.8 & 7.6  & 405  \\
Qwen3-14B & 66.5 & 28.9 & 565  \\
Qwen3-4B  & 64.0 & 9.5  & 538  \\
Qwen3-4B + SFT (Only output SQL, Short-Horizon) & 73.1 & 1.0  & 49   \\
Qwen3-4B + Ours (Max Output limitation = 2000 token) & 72.8 & 12.2 & 974  \\
Qwen3-4B + Ours (Max Output limitation = 4000 token) & 77.0 & 15.6 & 1266 \\
Qwen3-4B + Ours (Max Output limitation = 6000 token) & 78.8 & 19.1 & 1434 \\
Qwen3-4B + Ours (Max Output limitation = 8000 token) & 79.9 & 28.3 & 1880 \\
\bottomrule
\end{tabularx}
\caption{Efficiency-Performance Tradeoff: Latency-Performance Tradeoff. The results demonstrate that our method achieves the pareto frontier between efficiency and performance.}
\label{tab:latency_tokens}
\end{table*}

\begin{figure}[h]
  \centering
  \includegraphics[width=\linewidth]{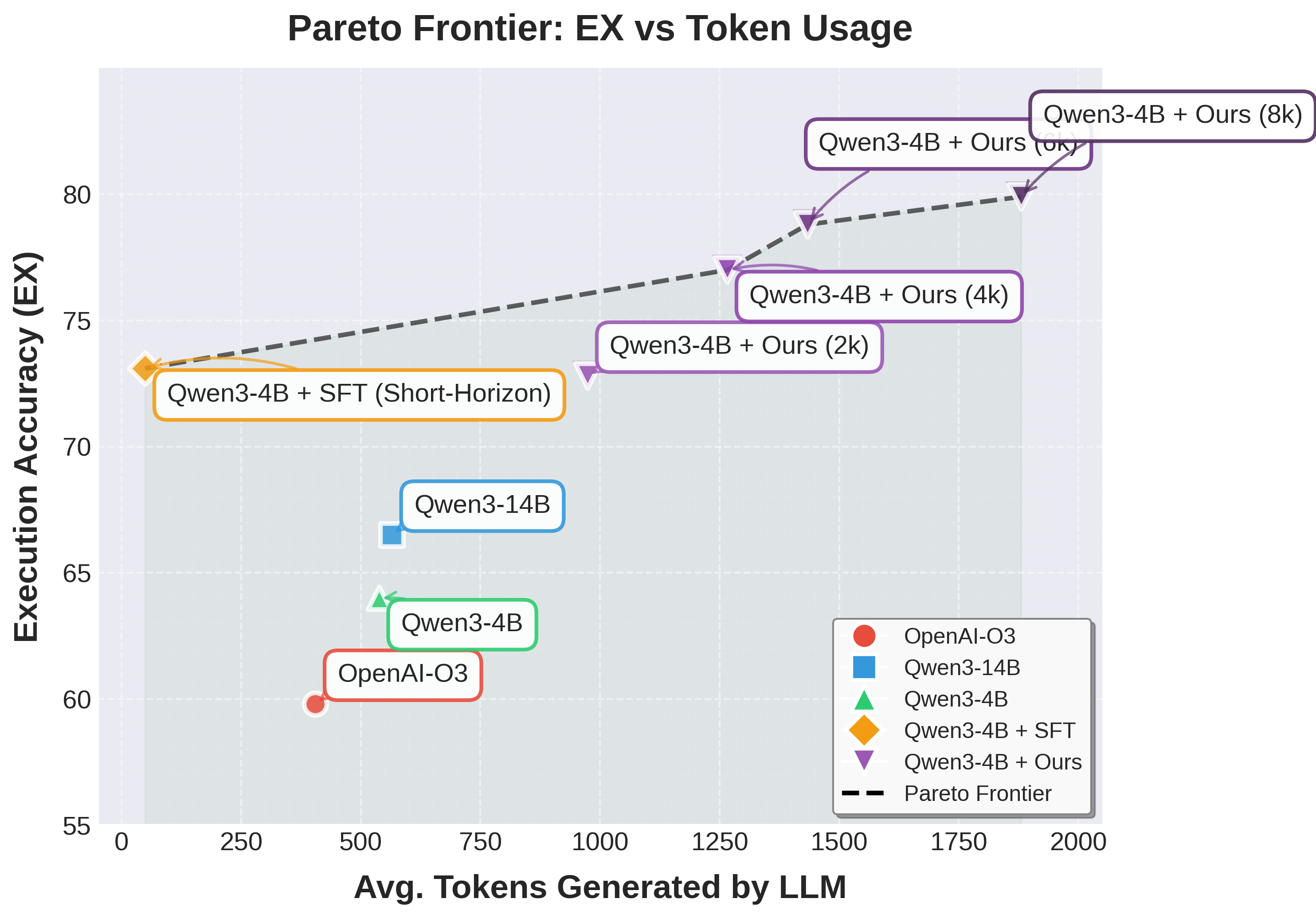}
  \caption{Pareto Frontier: Efficiency-Performance Tradeoff. The results demonstrate that our method achieves the Pareto frontier between inference tokens and performance.}
  \label{fig:pareto_frontier_ex_vs_tokens}
\end{figure}

\subsection{Quantifying Long-Horizon Abilities}

\begin{figure}[h]
  \centering
  \includegraphics[width=\linewidth]{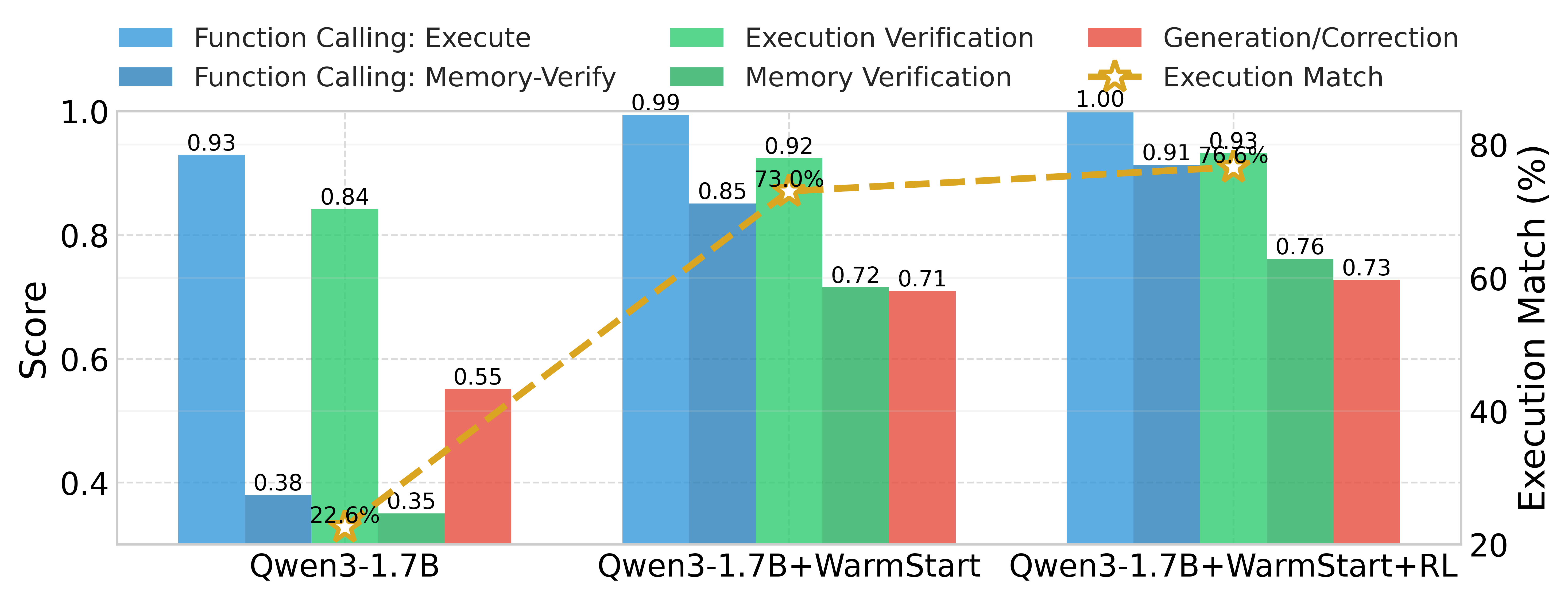}
  \caption{The evolution of different Long-Horizon Abilities and related Execution Match performance from base model to RL model for Qwen3-1.7B.}
  \label{fig:perf-17b}
\end{figure}

The evolution of Long-Horizon Abilities of Qwen3-1.7B model is shown in Fig.~\ref{fig:perf-17b}.

\subsection{RQ4: Training Dynamics}
\label{sec:RQ4TD}

\begin{figure}[h]
  \centering
  \includegraphics[width=\linewidth]{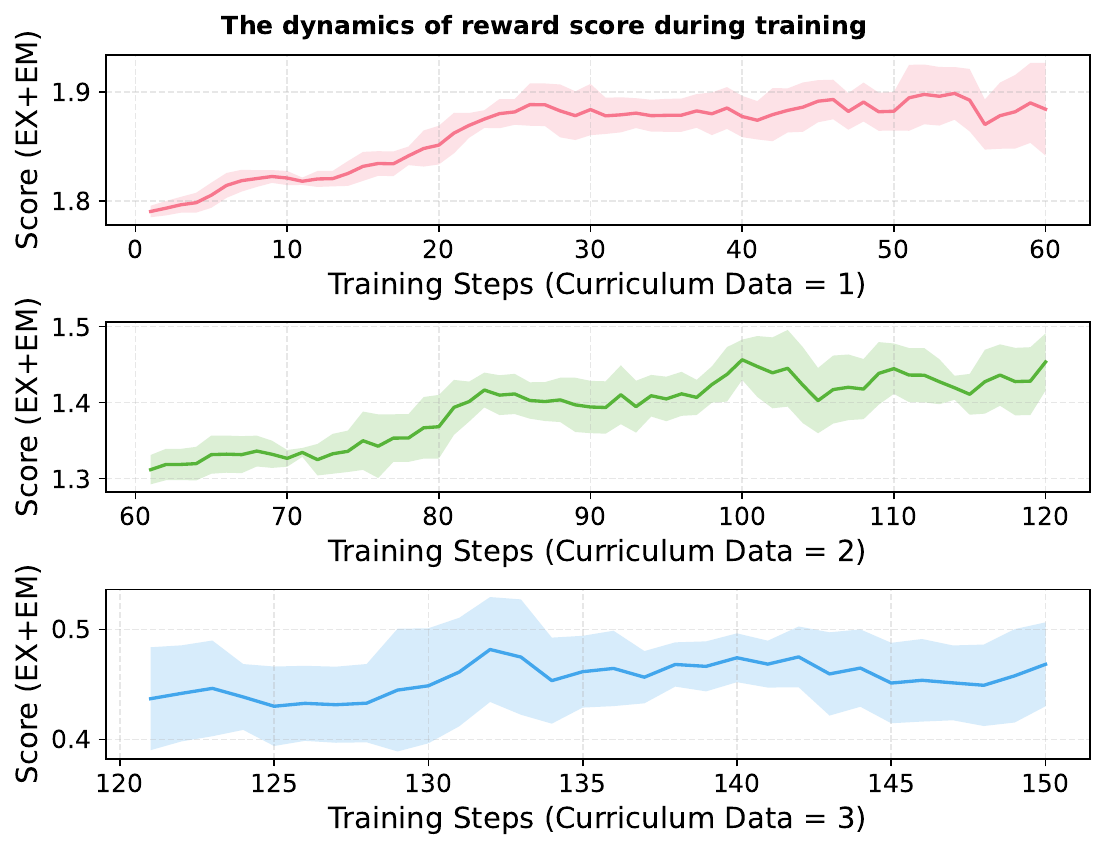} 
  \caption{The dynamics of reward score during outcome-reward based training.}
  \label{fig:train_reward_score}
\end{figure}

\begin{figure}[h]
  \centering
  \includegraphics[width=\linewidth]{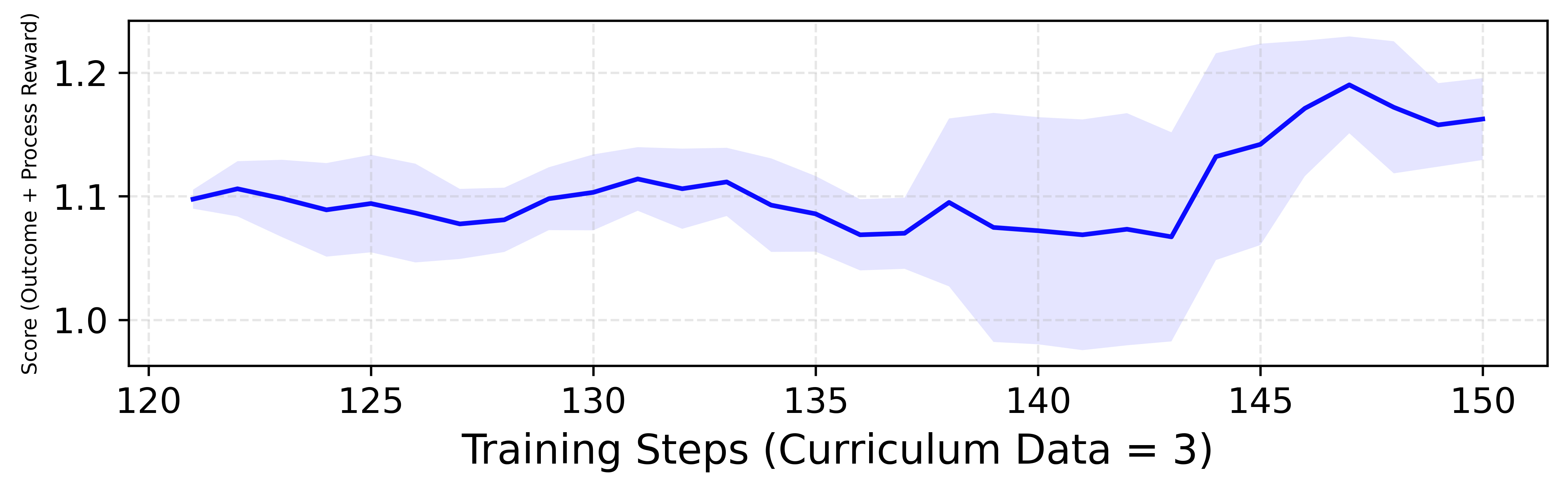} 
  \caption{The dynamics of reward score during outcome + process reward training for the last batch of curriculum data.}
  \label{fig:train_reward_score_process}
\end{figure}

\begin{figure}[h]
  \centering
  \includegraphics[width=\linewidth]{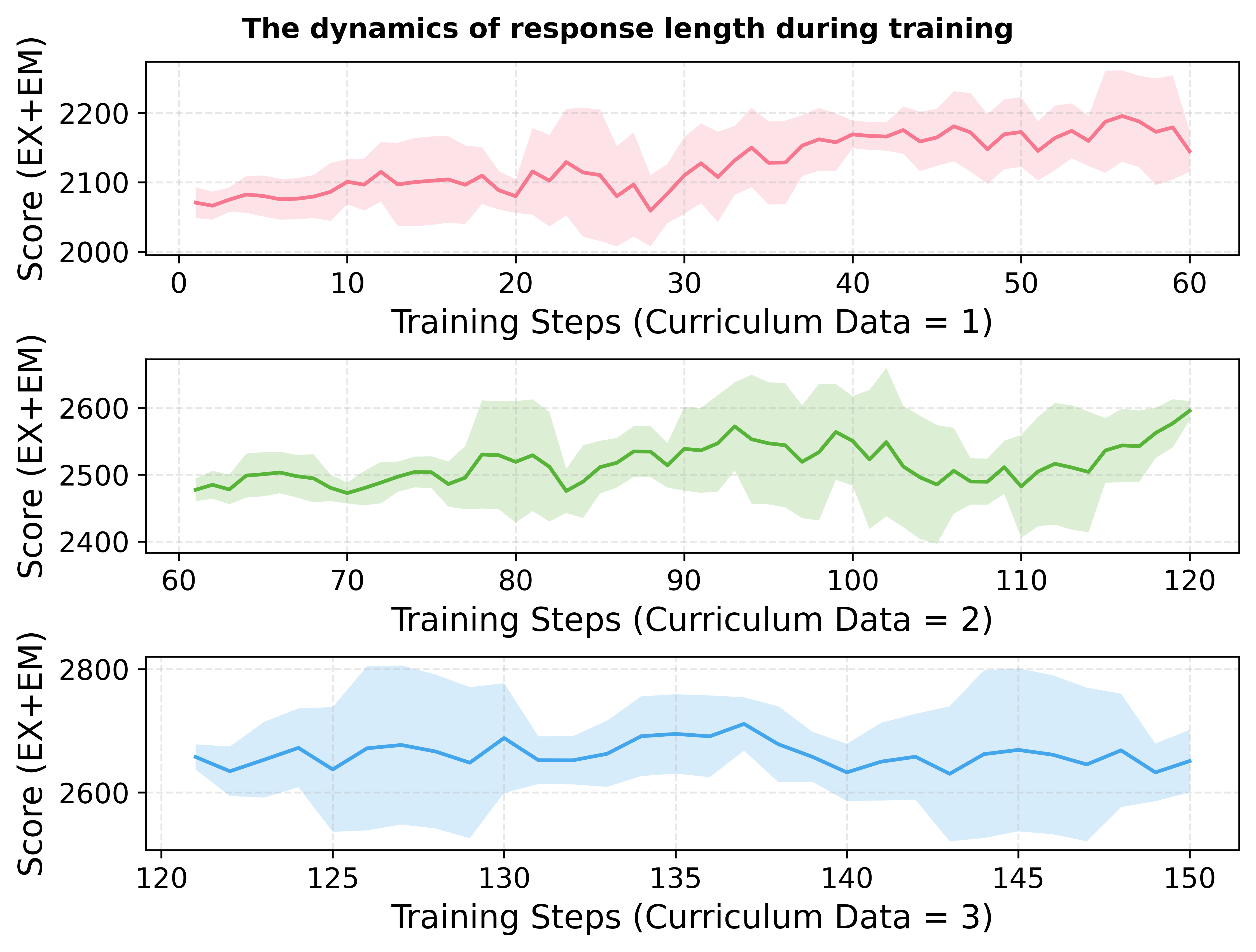} 
  \caption{The dynamics of response length during training.}
  \label{fig:length_score}
\end{figure}

\begin{figure}[H]
  \centering
  \includegraphics[width=\linewidth]{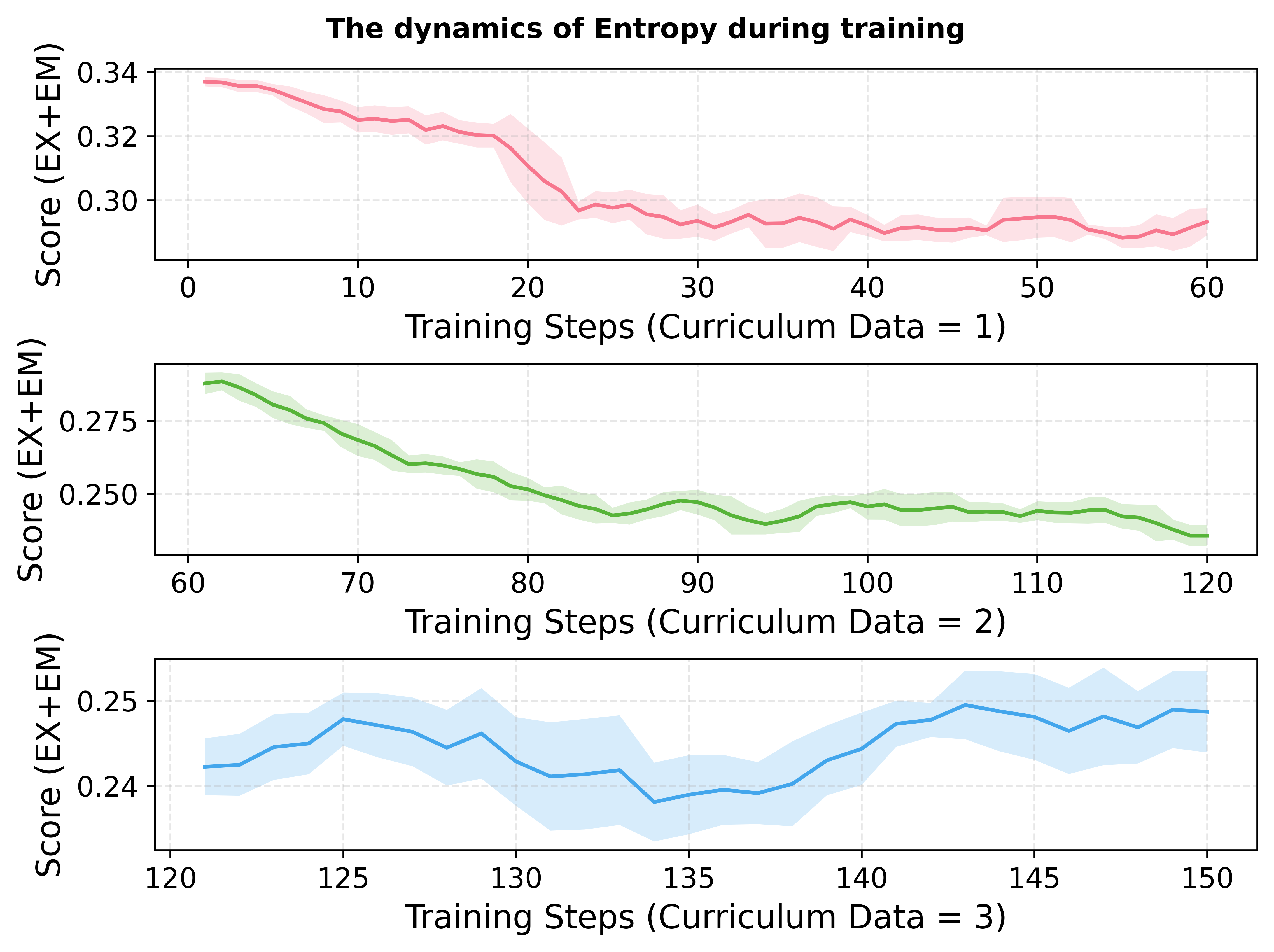} 
  \caption{The dynamics of entropy score during training.}
  \label{fig:train_entropy_score}
\end{figure}

Recall that we partition training samples by difficulty, estimated from the model's performance for curriculum RL training. We then examine the dynamics of reward, response length, and entropy. The reward is shown in Fig.~\ref{fig:train_reward_score}; entropy is shown in Fig.~\ref{fig:train_entropy_score}; and response length is shown in Fig.~\ref{fig:length_score}. We observe: (1) For curriculum levels $=1$ and $=2$ (easy/medium samples), the reward rises rapidly, whereas for level $=3$ (hard samples) it increases more gradually, indicating the model learns more slowly on difficult cases. The combined outcome+process reward is relatively smooth but trends upward throughout as shown in Fig.~\ref{fig:train_reward_score_process}. (2) Response length exhibits a similar pattern, and entropy drops sharply early on before stabilizing at a lower level.

\begin{figure}[htbp]
  \centering
  \includegraphics[width=\linewidth]{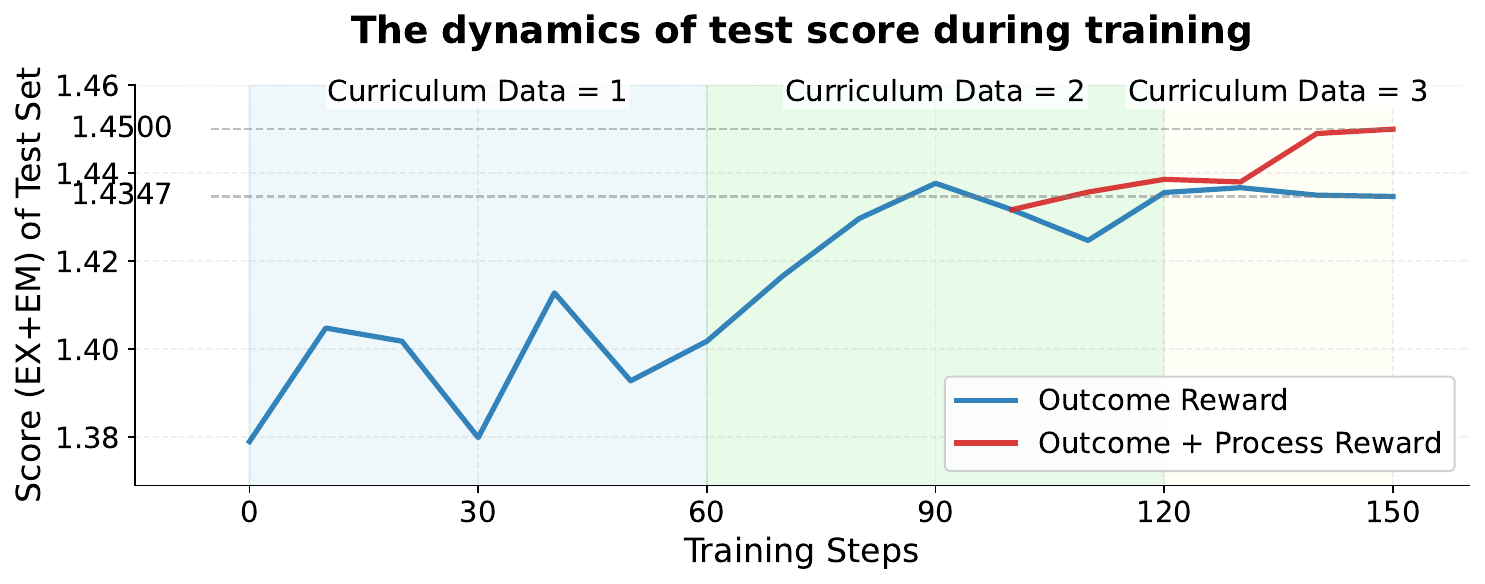} 
  \caption{The dynamics of the test score for different training checkpoints.}
  \label{fig:test_score_reward}
\end{figure}

Building on the training metrics above, we next track test-set scores over the course of training. As shown in Fig.~\ref{fig:test_score_reward}, using curriculum levels $1$ and $2$ yields substantial test-set gains early on. In later phases, as samples become harder, outcome rewards are sparser and improvements plateau. Incorporating dense and process rewards provides more frequent learning signals than outcome-only rewards, helping the model continue improving when outcome feedback alone is insufficient.

\section{Comparison between the Short-Horizon Reasoning Models (Qwen4B) and the Long-Horizon Reasoning Given the same question}
\label{sec:appendix}

\onecolumn

\subsection{Case 1: Base Model Fails but Long-Horizon Reasoning Model Succeeds with the help of ``Execution''-related Action}
The Difficulty of this case: Medium; The turn level is Turn 2.

\begin{prompt}
You are a SQL expert. You are given a question and you need to translate it to SQL step by step. 
Reasoning step by step. Once you feel you are ready for the final SQL, directly return the SQL 
inside <answer_sql> and </answer_sql>  at the end of your response.
Here are previous question and corresponding correct SQL in this dialogue: 

## Turn 1 ## 
Database schema: 
create table continents (
    ContId number,
    Continent text,
    primary key (ContId)
)
/*
1 example rows from table continents:
ContId	Continent
1	america
*/
create table countries (
    CountryId number,
    CountryName text,
    Continent number,
    primary key (CountryId),
    foreign key (Continent) references continents(ContId)
)
/*
1 example rows from table countries:
CountryId	CountryName	Continent
1	usa	1
*/
create table car_makers (
    Id number,
    Maker text,
    FullName text,
    Country text,
    primary key (Id),
    foreign key (Country) references countries(CountryId)
)
/*
1 example rows from table car_makers:
Id	Maker	FullName	Country
1	amc	American Motor Company	1
*/
create table model_list (
    ModelId number,
    Maker number,
    Model text,
    primary key (ModelId),
    foreign key (Maker) references car_makers(Id)
)
/*
1 example rows from table model_list:
ModelId	Maker	Model
1	1	amc
*/
create table car_names (
    MakeId number,
    Model text,
    Make text,
    primary key (MakeId),
    foreign key (Model) references model_list(Model)
)
/*
1 example rows from table car_names:
MakeId	Model	Make
1	chevrolet	chevrolet chevelle malibu
*/
create table cars_data (
    Id number,
    MPG text,
    Cylinders number,
    Edispl number,
    Horsepower text,
    Weight number,
    Accelerate number,
    Year number,
    primary key (Id),
    foreign key (Id) references car_names(MakeId)
)
/*
1 example rows from table cars_data:
Id	MPG	Cylinders	Edispl	Horsepower	Weight	Accelerate	Year
1	18	8	307.0	130	3504	12.0	1970
*/
Question: Can you list the number of car makers on each continent? " 
Corresponding Correct SQL: "SELECT T1.Continent ,  count ( * )  FROM CONTINENTS AS T1 JOIN COUNTRIES AS T2 ON T1.ContId   =   T2.continent JOIN car_makers AS T3 ON T2.CountryId   =   T3.Country GROUP BY T1.Continent" 

Now please translate the following question to SQL step by step 
Question: What about the number of car makers in the country USA? (Note you only need to translate the question to SQL instead answer the question. Once you feel you are ready for the final SQL, directly return the SQL inside <answer_sql> and </answer_sql>  at the end of your response.)
\end{prompt}

\begin{respqwen}
<think>
Okay, let's tackle this problem. The user is asking for the number of car makers in the USA. First, I need to understand the database schema and the relationships between the tables.

The relevant tables here are continents, countries, car_makers, and possibly others. The question is about car makers in the USA, so I need to connect the countries table to the car_makers table through the Country foreign key.

First, I should find the ContId for the USA. The countries table has a CountryName column, and the continents table has Continent and ContId. So, I need to join countries with continents on the Continent column. Then, find the ContId where CountryName is 'USA'.

Once I have the ContId for the USA, I can look into the car_makers table and count how many entries have their Country set to the CountryId of the USA. Wait, the car_makers table has a Country column that references countries.CountryId. So, the steps are:

1. Join countries and continents to get the continent info for the USA.
2. Find the CountryId of the USA from the countries table.
3. Count the number of car_makers where Country equals that CountryId.

Alternatively, maybe I can directly join car_makers with countries on Country, then filter where CountryName is 'USA' and count the entries.

So the SQL would involve joining car_makers with countries on the Country column, then filtering where CountryName is 'USA', and then counting the number of rows.

Let me check the tables again. The car_makers.Country is a foreign key to countries.CountryId. So, the correct approach is to join car_makers with countries on car_makers.Country = countries.CountryId, then filter where countries.CountryName = 'USA', and count the number of car_makers.

So the SQL query would be:

SELECT COUNT(*) AS num_car_makers
FROM car_makers
JOIN countries ON car_makers.Country = countries.CountryId
WHERE countries.CountryName = 'USA';

That should give the count of car makers in the USA. Let me make sure there are no other steps. The user just wants the number, so this should be straightforward. The previous example used joins between continents, countries, and car_makers, but in this case, since we're specifically looking for USA, we can directly filter the countries table.
</think>

<answer_sql>
SELECT COUNT(*) AS num_car_makers
FROM car_makers
JOIN countries ON car_makers.Country = countries.CountryId
WHERE countries.CountryName = 'USA';
</answer_sql>
\end{respqwen}

\begin{promptours}
You are a SQL expert. Your task is to translate a natural language question into SQL through step-by-step reasoning. Please follow the steps:
1. Reasoning
- Always think step by step before calling the tool. Draft the SQL.
2. Calling `exec_sql` tool (Please call `exec_sql` tool at least once)
- Call the `exec_sql` tool to execute the current generated SQL and verify the execution results based on questions.
- conclude <exec_verify>pass</exec_verify> if results are reasonable, otherwise <exec_verify>no_pass</exec_verify>.
- If no_pass, refine the SQL using the execution results and repeat call `exec_sql` tool until it passes.

Note:
1. Please call `exec_sql` tool at least once
2. Return the final SQL enclosed in: <answer_sql> ... </answer_sql>
Here are previous question and corresponding correct SQL in this dialogue: 

## Turn 1 ## 
"Database schema: 
create table continents (
    ContId number,
    Continent text,
    primary key (ContId)
)
/*
1 example rows from table continents:
ContId	Continent
1	america
*/
create table countries (
    CountryId number,
    CountryName text,
    Continent number,
    primary key (CountryId),
    foreign key (Continent) references continents(ContId)
)
/*
1 example rows from table countries:
CountryId	CountryName	Continent
1	usa	1
*/
create table car_makers (
    Id number,
    Maker text,
    FullName text,
    Country text,
    primary key (Id),
    foreign key (Country) references countries(CountryId)
)
/*
1 example rows from table car_makers:
Id	Maker	FullName	Country
1	amc	American Motor Company	1
*/
create table model_list (
    ModelId number,
    Maker number,
    Model text,
    primary key (ModelId),
    foreign key (Maker) references car_makers(Id)
)
/*
1 example rows from table model_list:
ModelId	Maker	Model
1	1	amc
*/
create table car_names (
    MakeId number,
    Model text,
    Make text,
    primary key (MakeId),
    foreign key (Model) references model_list(Model)
)
/*
1 example rows from table car_names:
MakeId	Model	Make
1	chevrolet	chevrolet chevelle malibu
*/
create table cars_data (
    Id number,
    MPG text,
    Cylinders number,
    Edispl number,
    Horsepower text,
    Weight number,
    Accelerate number,
    Year number,
    primary key (Id),
    foreign key (Id) references car_names(MakeId)
)
/*
1 example rows from table cars_data:
Id	MPG	Cylinders	Edispl	Horsepower	Weight	Accelerate	Year
1	18	8	307.0	130	3504	12.0	1970
*/
Question: Can you list the number of car makers on each continent? " 
Corresponding Correct SQL: "SELECT T1.Continent ,  count ( * )  FROM CONTINENTS AS T1 JOIN COUNTRIES AS T2 ON T1.ContId   =   T2.continent JOIN car_makers AS T3 ON T2.CountryId   =   T3.Country GROUP BY T1.Continent" 

Now please translate the following question to SQL step by step 
Question: What about the number of car makers in the country USA? (Note you only need to translate the question to SQL instead answer the question. Once you feel you are ready for the final SQL, directly return the SQL inside <answer_sql> and </answer_sql>  at the end of your response. 
 Note please call `exec_sql` tool at least once)
\end{promptours}

\begin{StepBox}{Response From Ours}
  \begin{ThoughtBox}
<think>
Okay, let's see. The user is asking for the number of car makers in the USA. The previous question was about the number of car makers on each continent, and the SQL for that involved joining continents, countries, and car_makers tables.

So, for this new question, the USA is a specific country. The country's ID in the countries table is 1, based on the example data provided. But maybe I shouldn't assume that; perhaps the SQL should use the country name directly. Wait, the countries table has CountryName and CountryId. The car_makers table has a Country column that references countries.CountryId. So to find car makers in the USA, I need to find the CountryId for the USA in the countries table and then count the car_makers with that CountryId.

Alternatively, maybe I can join the countries table with the car_makers table on CountryId, and then filter where CountryName is 'USA'. But the example data shows that the USA's CountryId is 1. However, relying on the example data might not be safe. The correct approach would be to use the CountryName 'USA' in the WHERE clause.

Wait, the previous SQL used a join between continents and countries. For this question, since we're dealing with a specific country, maybe the SQL would be:

SELECT COUNT(*) FROM car_makers AS T3 JOIN countries AS T2 ON T3.Country = T2.CountryId WHERE T2.CountryName = 'USA';

But let me check the tables again. The car_makers table has a Country column that's a foreign key to countries.CountryId. So joining car_makers with countries on car_makers.Country = countries.CountryId. Then filter where countries.CountryName is 'USA'.

Yes, that makes sense. So the SQL would be selecting count(*) from car_makers joined with countries on the CountryId, where the CountryName is USA.

Alternatively, maybe the user wants to use the previous SQL structure but add a WHERE clause for the continent. Wait, the previous SQL grouped by continent. Now, since the USA is a country, perhaps the SQL should be similar but with a WHERE condition on the country.

Wait, the previous answer joined continents, countries, and car_makers. But for this question, maybe we can simplify it. Since the USA is a country, we can directly join countries and car_makers, filter by CountryName = 'USA', and count the car_makers.

So the SQL would be:

SELECT COUNT(*) FROM car_makers JOIN countries ON car_makers.Country = countries.CountryId WHERE countries.CountryName = 'USA';

But I need to make sure that the joins are correct. Let me check the foreign keys again. The car_makers.Country references countries.CountryId. So yes, joining on car_makers.Country = countries.CountryId is correct.

Alternatively, maybe the user wants to use the same approach as the previous query but filter for the USA. The previous query used a group by on Continent. Here, since it's a specific country, we don't need grouping, just a count.

So the SQL should be as above. Now, I need to call exec_sql to verify. Let me generate that SQL and check if it's correct.
</think>
  \end{ThoughtBox}

  \begin{ActionBox}
<tool_call>
{"name": "exec_sql", "arguments": {"code": "SELECT COUNT(*) FROM car_makers JOIN countries ON car_makers.Country = countries.CountryId WHERE countries.CountryName = 'USA';"}}
</tool_call>
  \end{ActionBox}

  \begin{ObsBox}
<tool_response>

Recap:  
- Current question: Question: What about the number of car makers in the country USA?  
- Generated SQL: SELECT COUNT(*) FROM car_makers JOIN countries ON car_makers.Country = countries.CountryId WHERE countries.CountryName = 'USA';  
- SQL execution results (truncated to 200 characters): The sql results example is: [(0,)]  

Now please:  
1. Verify whether the SQL execution results are valid:  
   - Check if the SQL runs without errors.  
   - Check if the returned columns exist in the schema and are relevant to the question.  
   - Check if the results contain unexpected NULL values, empty sets, or error messages.  

2. After verifying, output:  
   - <exec_verify>pass</exec_verify> if the results are valid and consistent with the schema.  
   - <exec_verify>no_pass</exec_verify> if the results show errors, irrelevant columns, or invalid values.  

3. If <exec_verify>no_pass</exec_verify>, think step by step, refine the SQL and provide a corrected SQL and then execute it via re-calling ``exec_sql`` tool again via <tool_call>. Repeat until you get valid results.
4. If <exec_verify>pass</exec_verify>,  You have to call `memory_retrieve` tool via <tool_call>  at least once to ensure the current generated SQL is coherent with the historical memory. 
</tool_response>
  \end{ObsBox}
  
\begin{ThoughtBox}
<think>
(*@\textcolor{red}{Okay, let's see. The user asked for the number of car makers in the USA. The SQL I generated was joining car\_makers with countries on CountryId and filtering where CountryName is 'USA'. The execution result was [(0,)], which means zero rows. But looking at the example data provided earlier, there's a car\_makers entry with Country 1, which is the USA in the countries table. So why is the count zero? }@*)

(*@\textcolor{red}{Wait, maybe the CountryName in the countries table isn't exactly 'USA'. The example row shows CountryName as 'usa' with lowercase letters. The SQL uses 'USA' with uppercase letters. That's probably why it's not matching. The WHERE clause is case-sensitive, so 'usa' != 'USA'.}@*)

So I need to adjust the SQL to use 'usa' instead of 'USA' in the WHERE clause. Let me correct that and run the query again.
</think>
\end{ThoughtBox}

\begin{ActionBox}
<tool_call>
{"name": "exec_sql", "arguments": {"code": "SELECT COUNT(*) FROM car_makers JOIN countries ON car_makers.Country = countries.CountryId WHERE countries.CountryName = 'usa';"}}
</tool_call>
\end{ActionBox}

\begin{ObsBox}
<tool_response>
Recap:  
- Current question: Question: What about the number of car makers in the country USA?  
- Generated SQL: SELECT COUNT(*) FROM car_makers JOIN countries ON car_makers.Country = countries.CountryId WHERE countries.CountryName = 'usa';  
- SQL execution results (truncated to 200 characters): The sql results example is: [(4,)]  

Now please:  
1. Verify whether the SQL execution results are valid:  
   - Check if the SQL runs without errors.  
   - Check if the returned columns exist in the schema and are relevant to the question.  
   - Check if the results contain unexpected NULL values, empty sets, or error messages.  

2. After verifying, output:  
   - <exec_verify>pass</exec_verify> if the results are valid and consistent with the schema.  
   - <exec_verify>no_pass</exec_verify> if the results show errors, irrelevant columns, or invalid values.  

3. If <exec_verify>no_pass</exec_verify>, think step by step, refine the SQL and provide a corrected SQL and then execute it via re-calling ``exec_sql`` tool again via <tool_call>. Repeat until you get valid results.
4. If <exec_verify>pass</exec_verify>,  You have to call `memory_retrieve` tool via <tool_call>  at least once to ensure the current generated SQL is coherent with the historical memory. 
</tool_response>
\end{ObsBox}

\begin{ThoughtBox}
<think>
Okay, let's check the results. The SQL query now returns [(4,)], which means there are 4 car makers in the USA. Looking at the example data provided earlier, the car_makers table has an entry with Country 1, which corresponds to the USA in the countries table. The example data shows one entry, but the count is 4, which suggests there are more entries in the actual data that meet the criteria.

The SQL query is correct: it joins car_makers with countries on CountryId, filters for CountryName 'usa' (lowercase), and counts the rows. The result is a single row with the count 4, which is valid. There are no errors in the query, and the result makes sense given the data. Now, I need to call the memory_retrieve tool to ensure this SQL is coherent with the historical memory.
</think>
\end{ThoughtBox}

\begin{ActionBox}
<tool_call>
{"name": "memory_retrieve", "arguments": {"code": "SELECT COUNT(*) FROM car_makers JOIN countries ON car_makers.Country = countries.CountryId WHERE countries.CountryName = 'usa';"}}
</tool_call>
\end{ActionBox}

\begin{ObsBox}
<tool_response>

You are a coherence verifier for Multi-turn Text2SQL.

Current Question: Question: What about the number of car makers in the country USA?  
Proposed SQL: SELECT COUNT(*) FROM car_makers JOIN countries ON car_makers.Country = countries.CountryId WHERE countries.CountryName = 'usa';  
The execution results of the proposed SQL: The sql results example is: [(4,)]

Memory (historical information in order):  

== Turn 0 ==
Question: Question: Can you list the number of car makers on each continent?
Ground-Truth SQL: SELECT T1.Continent ,  count ( * )  FROM CONTINENTS AS T1 JOIN COUNTRIES AS T2 ON T1.ContId   =   T2.continent JOIN car_makers AS T3 ON T2.CountryId   =   T3.Country GROUP BY T1.Continent
Parsed Elements for each term: {'from': {'table_units': [('table_unit', '__continents__'), ('table_unit', '__countries__'), ('table_unit', '__car_makers__')], 'conds': [(False, 2, (0, (0, '__continents.contid__', False), None), (0, '__countries.continent__', False), None), 'and', (False, 2, (0, (0, '__countries.countryid__', False), None), (0, '__car_makers.country__', False), None)]}, 'select': ['__continents.continent__', '__all__'], 'where': [], 'groupBy': [(0, '__continents.continent__', False)], 'having': [], 'orderBy': [], 'limit': None, 'intersect': None, 'union': None, 'except': None}
SQL Results (truncated to 50 characters): [('america', 4), ('asia', 7), ('europe', 11)]
== Turn 0 ==

Your tasks:  
1. Verify whether the Proposed SQL is coherent with the Current Question and the Memory, based on the relation between the Current Question and Historical Questions.  
   - If the Current Question introduces changes (new columns, conditions, ordering, etc.), SQL should update accordingly.  
   - If not, SQL must remain consistent with the Historical Questions.  

Step-by-step reasoning checklist:  
   1. First parse the Proposed SQL into its components (SELECT, FROM, WHERE, GROUP BY, HAVING, ORDER BY, JOINs).
   2. Check tables are consistent with context.  
   3. Check selected columns match current and historical intent.  
   4. Check conditions (WHERE/GROUP/HAVING) reflect the relation between current and past questions.  
   5. Check ordering (ORDER BY) is preserved unless explicitly changed.  
   6. Verify that joins and table relationships follow the established context.
   7. Make sure the SQL and the execution results of the proposed SQL answer the current question while remaining logically coherent with the conversation history and execution results.

2. After verifying, output one of the following:  
   - `<memory_verify>pass</memory_verify>` if coherent.  
   - `<memory_verify>no_pass</memory_verify>` if not coherent.  

3. If `no_pass`: explain issues, think step by step to refine SQL, and you can call `exec_sql` tool again via <tool_call> to check the corrected SQL and get the execution results until you get `pass`. 
4. If `pass`: return the final SQL inside `<answer_sql>...</answer_sql>`.  

Note finally you should return the final SQL inside `<answer_sql>...</answer_sql>

</tool_response>
\end{ObsBox}

\begin{ThoughtBox}
<think>
Okay, let's verify the coherence of the proposed SQL with the current question and historical memory.

1. **Parse the Proposed SQL**: The SQL selects the count of car makers where the country name is 'usa'. It joins car_makers with countries on CountryId and filters by CountryName.

2. **Tables Consistency**: The tables involved (car_makers, countries) are correct based on the schema. The join condition (car_makers.Country = countries.CountryId) is accurate as per the foreign key relationship.

3. **Selected Columns**: The SELECT clause uses COUNT(*), which is appropriate for counting the number of car makers. The WHERE clause filters for 'usa' in CountryName, which aligns with the current question about the USA.

4. **Conditions**: The WHERE clause correctly targets the USA by CountryName. The historical question grouped by continent, but this question is specific to a country, so the condition is appropriate.

5. **Ordering**: There's no ORDER BY in the SQL, which is fine since the question just needs a count, not sorted results.

6. **Joins and Relationships**: The join between car_makers and countries is correct. The historical SQL also used similar joins, so this is consistent.

7. **Execution Results**: The result [(4,)] indicates 4 car makers in the USA, which matches the example data's implication (the sample car_makers entry with Country 1, which is the USA).

The SQL is coherent with the current question and historical context. It correctly adjusts the historical query from grouping by continent to filtering by a specific country. The execution result is valid.
</think>

<answer_sql>SELECT COUNT(*) FROM car_makers JOIN countries ON car_makers.Country = countries.CountryId WHERE countries.CountryName = 'usa';</answer_sql>
\end{ThoughtBox}
\end{StepBox}

\subsection{Case 2: Base Model Fails but Long-Horizon Reasoning Model Succeeds with the help of ``Memory-Verification''-related Action}

The Difficulty of this case: Medium; The turn level is Turn 2.

\begin{prompt}
You are a SQL expert. You are given a question and you need to translate it to SQL step by step.
Reasoning step by step before any tool call. You should use the text2sql tool after step by step solving the question,  before generate final answer at least once and refine your answer if necessary. 
Once you feel you are ready for the final SQL, directly return the SQL inside <answer_sql> and </answer_sql>  at the end of your response.
Here are previous question and corresponding correct SQL in this dialogue: 

## Turn 1 ## 
User: "Database schema: 
create table city (
    ID number,
    Name text,
    CountryCode text,
    District text,
    Population number,
    primary key (ID),
    foreign key (CountryCode) references country(Code)
)
/*
1 example rows from table city:
ID	Name	CountryCode	District	Population
1	Kabul	AFG	Kabol	1780000
*/
create table sqlite_sequence (
    name text,
    seq text
)
/*
1 example row from table sqlite_sequence:
name	seq
city	4079
*/
create table country (
    Code text,
    Name text,
    Continent text,
    Region text,
    SurfaceArea number,
    IndepYear number,
    Population number,
    LifeExpectancy number,
    GNP number,
    GNPOld number,
    LocalName text,
    GovernmentForm text,
    HeadOfState text,
    Capital number,
    Code2 text,
    primary key (Code)
)
/*
1 example rows from table country:
Code	Name	Continent	Region	SurfaceArea	IndepYear	Population	LifeExpectancy	GNP	GNPOld	LocalName	GovernmentForm	HeadOfState	Capital	Code2
ABW	Aruba	North America	Caribbean	193.0	None	103000	78.4	828.0	793.0	Aruba	Nonmetropolitan Territory of The Netherlands	Beatrix	129	AW
*/
create table countrylanguage (
    CountryCode text,
    Language text,
    IsOfficial text,
    Percentage number,
    primary key (CountryCode),
    foreign key (CountryCode) references country(Code)
)
/*
1 example rows from table countrylanguage:
CountryCode	Language	IsOfficial	Percentage
ABW	Dutch	T	5.3
*/
Question: How many government forms are in the table? " 
Corresponding Correct SQL: "SELECT count ( GovernmentForm )  FROM country" 

## Turn 2 ## 
User: "Question: How many of those have an average life expectancy that is longer than 72? " 
Corresponding Correct SQL: "SELECT GovernmentForm FROM country GROUP BY GovernmentForm HAVING avg ( LifeExpectancy )   >  72" 

Now please translate the following question to SQL step by step 
Question: What is the population of each of those government forms? (Note you only need to translate the question to SQL instead answer the question. Once you feel you are ready for the final SQL, directly return the SQL inside <answer_sql> and </answer_sql>  at the end of your response.)
\end{prompt}

\begin{respqwen}
<think>
Okay, let's tackle this question. The user is asking, "What is the population of each of those government forms?" So first, I need to understand what they mean. 

Looking back at previous interactions, the user mentioned "those" referring to the government forms from the country table. In the previous question, they asked about government forms and used a SQL query that grouped by GovernmentForm and applied an average life expectancy condition. Now, they want the population for each of those government forms.

So the main goal is to get the population for each distinct government form. The country table has the GovernmentForm and Population fields. 

First, I need to select the GovernmentForm and Population from the country table. But since there might be multiple entries per government form, I should group by GovernmentForm. However, the question is asking for the population of each form. Wait, but each government form might have multiple populations. Wait, no, each country has a government form, and each country's population is listed. So if there are multiple countries with the same government form, we need to aggregate their populations. But the question says "the population of each of those government forms" - does that mean the total population for each form? Or the population of each country that has that form?

Wait, the previous question was about government forms and average life expectancy. The user might be referring to the same government forms that were considered in that previous query. But the current question is about population. 

But the user's current question is not about filtering based on the previous condition. The note says to only translate the question to SQL, not answer it. So the question is: for each government form, what is the population. So perhaps they want the population of each country grouped by their government form. But the wording is a bit ambiguous. 

But the way the question is phrased, "the population of each of those government forms"—"those" refers to the government forms from the previous question. However, the previous question was about government forms that had an average life expectancy over 72. But the current question is not filtering based on that. Wait, the note says to only translate the question, not answer. So maybe the user is asking for the population of each government form, regardless of the previous condition. 

But the user's note says "Note you only need to translate the question to SQL instead answer the question." So the question is: "What is the population of each of those government forms?" where "those" refers to the government forms from the previous question. But the previous question was about government forms with average life expectancy over 72. However, the current question is not necessarily filtering based on that. Wait, but the user might be referring to the same "those" as in the previous question. But the note says to translate the question, not answer. So maybe the user is asking for the population of each government form, but the "those" is referring to the government forms that were in the previous query. 

But the user's current question is not explicitly filtering. However, the note says to translate the question as is. So perhaps the user is asking for the population of each government form, but the "those" is a reference to the government forms that were in the previous question. But the previous question was about government forms with average life expectancy over 72. However, the current question is not necessarily filtering. 

Alternatively, maybe the user is asking for the population of each government form, without any filtering. But the wording is ambiguous. However, the note says to translate the question to SQL, not answer. So the correct approach is to take the question as is. 

The question is: "What is the population of each of those government forms?" So "those" refers to the government forms from the previous question. But the previous question was about government forms that had an average life expectancy over 72. However, the current question is not necessarily filtering. But the user might be referring to the same government forms. 

But the note says to translate the question, not answer. So perhaps the user is asking for the population of each government form, but the "those" is a reference to the government forms that were in the previous question. However, the current question is not explicitly filtering. 

Alternatively, maybe the user is asking for the population of each government form, regardless of any previous conditions. 

But the user's note says to translate the question as is. So the correct SQL would be to select the population for each government form. But how? 

If the user wants the population of each government form, then we need to group by GovernmentForm and aggregate the population. But the question is phrased as "the population of each of those government forms"—so maybe they want the total population for each government form. 

So the SQL would be: SELECT GovernmentForm, SUM(Population) AS TotalPopulation FROM country GROUP BY GovernmentForm. 

But the user's question is not clear. However, the previous question was about government forms with average life expectancy over 72. But the current question is not filtering. 

Alternatively, maybe the user is asking for the population of each country grouped by their government form. But that would be selecting GovernmentForm and Population, but that would give multiple populations per government form. 

But the question says "the population of each of those government forms"—so perhaps they want the total population for each government form. 

So the correct SQL would be to group by GovernmentForm and sum the population. 

But the user's note says to translate the question, not answer. So the correct SQL would be:

SELECT GovernmentForm, SUM(Population) AS Population FROM country GROUP BY GovernmentForm;

But the user might be expecting to get the population of each country, but that's not possible because each country has a government form. However, the question is ambiguous. 

Alternatively, maybe the user is asking for the population of each government form, meaning the population of each country that has that form. But that would be a list of populations per government form. 

But the question is not clear. However, based on the previous interaction, the user might be referring to the same "those" as in the previous question. The previous question was about government forms with average life expectancy over 72. So the current question is asking for the population of each of those government forms. 

So the SQL would be: SELECT Population FROM country WHERE GovernmentForm IN (SELECT GovernmentForm FROM country GROUP BY GovernmentForm HAVING AVG(LifeExpectancy) > 72) GROUP BY GovernmentForm;

But the user's note says to translate the question, not answer. So the correct approach is to take the question as is. 

But the user's question is "What is the population of each of those government forms?" where "those" refers to the government forms from the previous question. So the SQL would be to first get the government forms from the previous query (those with average life expectancy over 72), then get the population for each of those forms. 

But the user's note says to translate the question, not answer. So the correct SQL would be:

SELECT GovernmentForm, Population FROM country WHERE GovernmentForm IN (SELECT GovernmentForm FROM country GROUP BY GovernmentForm HAVING AVG(LifeExpectancy) > 72) GROUP BY GovernmentForm;

But the user might not need the IN clause. However, the question is about the population of each of those government forms, which were from the previous question. 

Alternatively, the user might be asking for the population of each government form, without any filtering. But the note says to translate the question as is. 

But the user's question is ambiguous. However, based on the previous interaction, the user is likely referring to the same "those" as in the previous question. Therefore, the correct SQL would be to select the population for each government form that was in the previous query. 

So the SQL would be:

SELECT GovernmentForm, Population FROM country WHERE GovernmentForm IN (SELECT GovernmentForm FROM country GROUP BY GovernmentForm HAVING AVG(LifeExpectancy) > 72) GROUP BY GovernmentForm;

But the user might not need the GROUP BY here. Wait, the question is "What is the population of each of those government forms?" So for each government form, what is the population. If there are multiple countries with the same government form, the population would be the sum. But the question is not clear. 

Alternatively, the user might want to list the population of each country, grouped by their government form. But that would be multiple populations per form. 

But the question is not clear. However, the correct approach is to translate the question as per the user's instruction. The user's question is "What is the population of each of those government forms?" where "those" refers to the government forms from the previous question. 

Therefore, the SQL would be to first get the government forms from the previous query (those with average life expectancy over 72), then get the population for each of those forms. 

So the SQL would be:

SELECT GovernmentForm, Population FROM country WHERE GovernmentForm IN (SELECT GovernmentForm FROM country GROUP BY GovernmentForm HAVING AVG(LifeExpectancy) > 72) GROUP BY GovernmentForm;
</think>

\end{respqwen}

\begin{promptours}
You are a SQL expert. Your task is to translate a natural language question into SQL through step-by-step reasoning. Please follow the steps:

1. Reasoning
- Always think step by step before calling the tool. Draft the SQL.

2. Calling `exec_sql` tool (Please call `exec_sql` tool at least once)
- Call the `exec_sql` tool to execute the current generated SQL and verify the execution results based on questions.
- conclude <exec_verify>pass</exec_verify> if results are reasonable, otherwise <exec_verify>no_pass</exec_verify>.
- If no_pass, refine the SQL using the execution results and repeat call `exec_sql` tool until it passes.

Note:
1. Please call `exec_sql` tool at least once
2. Return the final SQL enclosed in: <answer_sql> ... </answer_sql>

Here are previous question and corresponding correct SQL in this dialogue: 

## Turn 1 ## 
User: "Database schema: 
create table city (
    ID number,
    Name text,
    CountryCode text,
    District text,
    Population number,
    primary key (ID),
    foreign key (CountryCode) references country(Code)
)
/*
1 example rows from table city:
ID	Name	CountryCode	District	Population
1	Kabul	AFG	Kabol	1780000
*/
create table sqlite_sequence (
    name text,
    seq text
)
/*
1 example row from table sqlite_sequence:
name	seq
city	4079
*/
create table country (
    Code text,
    Name text,
    Continent text,
    Region text,
    SurfaceArea number,
    IndepYear number,
    Population number,
    LifeExpectancy number,
    GNP number,
    GNPOld number,
    LocalName text,
    GovernmentForm text,
    HeadOfState text,
    Capital number,
    Code2 text,
    primary key (Code)
)
/*
1 example rows from table country:
Code	Name	Continent	Region	SurfaceArea	IndepYear	Population	LifeExpectancy	GNP	GNPOld	LocalName	GovernmentForm	HeadOfState	Capital	Code2
ABW	Aruba	North America	Caribbean	193.0	None	103000	78.4	828.0	793.0	Aruba	Nonmetropolitan Territory of The Netherlands	Beatrix	129	AW
*/
create table countrylanguage (
    CountryCode text,
    Language text,
    IsOfficial text,
    Percentage number,
    primary key (CountryCode),
    foreign key (CountryCode) references country(Code)
)
/*
1 example rows from table countrylanguage:
CountryCode	Language	IsOfficial	Percentage
ABW	Dutch	T	5.3
*/
Question: How many government forms are in the table? " 
Corresponding Correct SQL: "SELECT count ( GovernmentForm )  FROM country" 

## Turn 2 ## 
User: "Question: How many of those have an average life expectancy that is longer than 72? " 
Corresponding Correct SQL: "SELECT GovernmentForm FROM country GROUP BY GovernmentForm HAVING avg ( LifeExpectancy )   >  72" 

Now please translate the following question to SQL step by step 
Question: What is the population of each of those government forms? (Note you only need to translate the question to SQL instead answer the question. Once you feel you are ready for the final SQL, directly return the SQL inside <answer_sql> and </answer_sql>  at the end of your response. 
 Note please call `exec_sql` tool at least once )
\end{promptours}

\begin{StepBox}{Response From Ours}
    \begin{ThoughtBox}
<think>
Okay, let's tackle this question step by step. The user is asking for the population of each government form. From the previous turns, I can see that the country table has the GovernmentForm and Population columns.

First, I need to group the countries by their GovernmentForm. Then, for each group, I should calculate the population. Wait, but the question says "the population of each of those government forms" - does that mean the total population per government form or the average? The previous question was about average life expectancy, so maybe here it's the total population. But the user didn't specify average, so maybe sum?

...

But given that the previous question was about government forms with average life >72, and this question is about their population, the correct SQL would be the one with the subquery. However, without explicit instructions, it's hard to be certain. But the user's instruction says to translate the question to SQL, not to answer it. So, the SQL should be based on the question's wording, not on previous context. Therefore, the SQL would be:

SELECT GovernmentForm, SUM(Population) FROM country GROUP BY GovernmentForm;

But the user might have intended to refer to the previous context. However, since the user hasn't specified that, it's safer to assume that "those" refers to all government forms. Therefore, the SQL is as above.

Now, I need to call the exec_sql tool with this SQL to verify. Let's proceed.
    \end{ThoughtBox}

\begin{ActionBox}
<tool_call>
{"name": "exec_sql", "arguments": {"code": "SELECT GovernmentForm, SUM(Population) FROM country GROUP BY GovernmentForm"}}
</tool_call>
\end{ActionBox}

\begin{ObsBox}
<tool_response>

Recap:  
- Current question: Question: What is the population of each of those government forms?  
- Generated SQL: SELECT GovernmentForm, SUM(Population) FROM country GROUP BY GovernmentForm  
- SQL execution results (truncated to 200 characters): The sql results example is: [('Administrated by the UN', 885000), ('Autonomous Area', 3101000), ('Co-administrated', 0), ('Commonwealth of the US', 3947000), ('Constitutional Monarchy', 405893000), ('  

Now please:  
1. Verify whether the SQL execution results are valid:  
   - Check if the SQL runs without errors.  
   - Check if the returned columns exist in the schema and are relevant to the question.  
   - Check if the results contain unexpected NULL values, empty sets, or error messages.  

2. After verifying, output:  
   - <exec_verify>pass</exec_verify> if the results are valid and consistent with the schema.  
   - <exec_verify>no_pass</exec_verify> if the results show errors, irrelevant columns, or invalid values.  

3. If <exec_verify>no_pass</exec_verify>, think step by step, refine the SQL and provide a corrected SQL and then execute it via re-calling ``exec_sql`` tool again via <tool_call>. Repeat until you get valid results.
4. If <exec_verify>pass</exec_verify>,  You have to call `memory_retrieve` tool via <tool_call>  at least once to ensure the current generated SQL is coherent with the historical memory. 
</tool_response>
\end{ObsBox}

\begin{ThoughtBox}
<think>
Okay, let's start by verifying the SQL execution results. The generated SQL is:

SELECT GovernmentForm, SUM(Population) FROM country GROUP BY GovernmentForm

The execution results show entries like ('Administrated by the UN', 885000), ('Autonomous Area', 3101000), etc. These seem to be valid government forms with their corresponding summed population values. The columns GovernmentForm and Population exist in the country table, so the SQL is correctly referencing existing columns. The results don't show any errors or unexpected NULL values. The SUM(Population) makes sense as it aggregates the population for each government form.

Since the results are valid and consistent with the schema, I'll mark it as pass. Next, I need to call the memory_retrieve tool to ensure coherence with historical memory. The previous correct SQLs were about counting government forms and averaging life expectancy. The current SQL is about summing population per government form, which is a logical next step and coherent with the previous queries. So, I'll call memory_retrieve with the current SQL.
</think>
\end{ThoughtBox}

\begin{ActionBox}
<tool_call>
{"name": "memory_retrieve", "arguments": {"code": "SELECT GovernmentForm, SUM(Population) FROM country GROUP BY GovernmentForm"}}
</tool_call>
\end{ActionBox}

\begin{ObsBox}
<tool_response>

You are a coherence verifier for Multi-turn Text2SQL.

Current Question: Question: What is the population of each of those government forms?  
Proposed SQL: SELECT GovernmentForm, SUM(Population) FROM country GROUP BY GovernmentForm  
The execution results of the proposed SQL: The sql results example is: [('Administrated by the UN', 885000), ('Autonomous Area', 3101000), ('Co-administrated', 0), ('Commonwealth of the US', 3947000), ('Constitutional Monarchy', 405893000), ('

Memory (historical information in order):  

== Turn 0 ==
Question: Question: How many government forms are in the table?
Ground-Truth SQL: SELECT count ( GovernmentForm )  FROM country
Parsed Elements for each term: {'from': {'table_units': [('table_unit', '__country__')], 'conds': []}, 'select': ['__country.governmentform__'], 'where': [], 'groupBy': [], 'having': [], 'orderBy': [], 'limit': None, 'intersect': None, 'union': None, 'except': None}
SQL Results (truncated to 50 characters): [(239,)]
== Turn 0 ==
    
== Turn 1 ==
Question: Question: How many of those have an average life expectancy that is longer than 72?
Ground-Truth SQL: SELECT GovernmentForm FROM country GROUP BY GovernmentForm HAVING avg ( LifeExpectancy )   >  72
Parsed Elements for each term: {'from': {'table_units': [('table_unit', '__country__')], 'conds': []}, 'select': ['__country.governmentform__'], 'where': [], 'groupBy': [(0, '__country.governmentform__', False)], 'having': [(False, 3, (0, (5, '__country.lifeexpectancy__', False), None), 72.0, None)], 'orderBy': [], 'limit': None, 'intersect': None, 'union': None, 'except': None}
SQL Results (truncated to 50 characters): [('Commonwealth of the US',), ('Constitutional Monarchy (Emirate)',), ('Constitutional Monarchy, Federation',), ('Dependent Territory of the UK',), ('Emirate Federation',), ('Federation',), ('Monarchy (Emirate)',), ('Monarchy (Sultanate)',), ('Nonmetropolitan Territory of France',), ('Nonmetropolitan Territory of The Netherlands',), ('Overseas Department of France',), ('Parliamentary Coprincipality',), ('Part of Denmark',), ('Socialistic Republic',), ('Socialistic State',), ('Special Administrative Region of China',), ('US Territory',)]
== Turn 1 ==

Your tasks:  
1. Verify whether the Proposed SQL is coherent with the Current Question and the Memory, based on the relation between the Current Question and Historical Questions.  
   - If the Current Question introduces changes (new columns, conditions, ordering, etc.), SQL should update accordingly.  
   - If not, SQL must remain consistent with the Historical Questions.  

Step-by-step reasoning checklist:  
   1. First parse the Proposed SQL into its components (SELECT, FROM, WHERE, GROUP BY, HAVING, ORDER BY, JOINs).
   2. Check tables are consistent with context.  
   3. Check selected columns match current and historical intent.  
   4. Check conditions (WHERE/GROUP/HAVING) reflect the relation between current and past questions.  
   5. Check ordering (ORDER BY) is preserved unless explicitly changed.  
   6. Verify that joins and table relationships follow the established context.
   7. Make sure the SQL and the execution results of the proposed SQL answer the current question while remaining logically coherent with the conversation history and execution results.

2. After verifying, output one of the following:  
   - `<memory_verify>pass</memory_verify>` if coherent.  
   - `<memory_verify>no_pass</memory_verify>` if not coherent.  

3. If `no_pass`: explain issues, think step by step to refine SQL, and then you can call `exec_sql` tool again via <tool_call> to check the corrected SQL and get the execution results  until you get `pass`. 
4. If `pass`: return the final SQL inside `<answer_sql>...</answer_sql>`.  

Note finally you should return the final SQL inside `<answer_sql>...</answer_sql>

</tool_response>
\end{ObsBox}

\begin{ThoughtBox}
<think>
Okay, let's start by breaking down the current question and the proposed SQL. The user is asking for the population of each of those government forms. The "those" likely refers to the government forms identified in the previous question, which were those with an average life expectancy longer than 72. 

The proposed SQL is selecting GovernmentForm and the sum of Population from the country table, grouped by GovernmentForm. This makes sense because it's aggregating the population for each government form. However, the previous question filtered government forms based on average life expectancy. The current SQL doesn't include that filter, so it's considering all government forms, not just those from the previous query.

(*@\textcolor{red}{Wait, but the user's question says "those government forms," which might refer to the ones from the previous question. The previous SQL used a HAVING clause to filter government forms with avg LifeExpectancy >72. The current SQL doesn't have that condition. That's a problem because the current question is about the population of each of those (previous) government forms. So the SQL should include the same HAVING clause as the previous query to ensure we're only considering those forms.}@*)

The execution results show entries like 'Administrated by the UN' and 'Autonomous Area', which might not be the ones filtered in the previous step. The previous query's results included forms like 'Commonwealth of the US' and 'Constitutional Monarchy'. The current SQL is summing populations for all government forms, not just the filtered ones. That's why the results include forms not in the previous HAVING clause.

So the error is that the current SQL doesn't include the HAVING condition from the previous query. To fix this, the SQL should include the same HAVING clause (avg(LifeExpectancy) >72) in the GROUP BY section. That way, it's only considering the government forms that met the previous condition when calculating their populations.  The generated SQL is: SELECT GovernmentForm, SUM(Population) FROM country GROUP BY GovernmentForm HAVING avg(LifeExpectancy) > 72 </think>
<answer_sql> SELECT GovernmentForm, SUM(Population) FROM country GROUP BY GovernmentForm HAVING avg(LifeExpectancy) > 72 </answer_sql>
\end{ThoughtBox}
    
\end{StepBox}

\phantomsection
\label{pg:Case2}

\section{Prompt for LLM-as-Judge Used to Classify SQL Errors}

  \begin{LLMjudgeBox}
You are an LLM-as-judge for **multi-turn Text2SQL**.
Given a dialogue history, the current gold SQL, and the model's predicted SQL,
output exactly ONE of the categories below (mutually exclusive Top-1), focusing ONLY on
**multi-turn coherence** failure modes.

Classify into a coherence-related category whenever possible; use "Others" only if the
error cannot reasonably be attributed to multi-turn incoherence.

Allowed `category` values:
  - "SchemaLinking (Coherence)"        # wrong tables/columns vs prior grounded focus
  - "JoinPath (Coherence)"             # tables OK, but relationship edges/bridge/keys diverge
  - "AggregationDrift (Coherence)"     # aggregated/ranked/window context lost or mutated
  - "ConstraintCoherence (Coherence)"  # constraint/value/scope incoherence (dropped/over-carry/scope/coref→value)
  - "Others"                           # correct, or error not plausibly due to multi-turn incoherence

### STRICT DEFINITIONS

1) "SchemaLinking (Coherence)"
   Prediction chooses the wrong **tables/columns** relative to previously grounded schema.
   Prior turns established certain tables (T_prev) or salient columns (C_prev) as the focus; the current SQL
   omits or swaps them despite **continuation cues** (“also”, “those”, “same”, “among those”, “of the above”).
   (Note: edges/joins belong to JoinPath, not here.)

2) "JoinPath (Coherence)"
   The conversation already established a **relationship chain** (edges/bridge tables/keys).
   The prediction uses a different/missing bridge or wrong join keys, changing which entities are selected.
   (Nodes/tables match prior focus, but edges/joins differ.)

3) "AggregationDrift (Coherence)"
   Prior turns established an aggregated/ranked/windowed view (GROUP BY, HAVING, ORDER BY, window functions).
   The prediction **drops or mutates** that context under continuation cues (“those top teams”, “highest average”).
   This includes loss/change of GROUP BY / HAVING / ORDER / LIMIT / window that was salient previously.

4) "ConstraintCoherence (Coherence)"
   Any **constraint/value/scope** incoherence vs prior turns, including:
     - Dropped constraints (under-carry): previously applied filters (e.g., year > 2015, city = 'Boston') vanish under continuation.
     - Over-carry (unwarranted carry): previous filters are kept despite a reset cue (“now overall”, “regardless”).
     - Result-set scope mismatch: should operate on the **subset from the previous result**, but queries the whole DB.
     - Coreference/Ellipsis → value/constraint mismatch: pronouns/ellipsis (“them/these/that city/same dept”)
       resolved to wrong literals/IDs, altering constraints vs prior context.

5) "Others"
   Use when: (a) the prediction is correct; (b) the error is not plausibly due to cross-turn incoherence;
   or (c) information is insufficient to attribute the error to (1) to (4).

### TIE-BREAK RULES (apply top-down; prefer coherence categories before "Others")
1) If the table/column set is wrong vs prior-grounded context → "SchemaLinking (Coherence)".
2) Else if tables are right but relationship edges/bridge/keys diverge → "JoinPath (Coherence)".
3) Else if aggregated/ranked/window context from prior is lost/mutated → "AggregationDrift (Coherence)".
4) Else if constraint/value/scope coherence is broken → "ConstraintCoherence (Coherence)".
5) Else → "Others".

### OUTPUT FORMAT (valid JSON only)
{{
  "category": "<one of: {', '.join(CATEGORIES)}>",
  "rationale": "2 to 4 sentences citing cross-turn evidence for the chosen category.",
  "cross_turn_signals": ["brief bullets of evidence"],
  "confidence": 0.0
}}
Keep the rationale concise and evidence-driven. No extra text outside the JSON.
"""
  \end{LLMjudgeBox}